\documentclass[10pt,twocolumn,letterpaper]{article}

\ifdefined\pdfinfoomitdate\pdfinfoomitdate=1\fi
\ifdefined\pdfsuppressptexinfo\pdfsuppressptexinfo=15\fi
\ifdefined\pdftrailerid\pdftrailerid{}\fi

\usepackage[pagenumbers]{cvpr}

\usepackage{algorithm}
\usepackage{algpseudocode}
\usepackage{colortbl}
\usepackage{tabularx}
\usepackage{arxivfloats}
\usepackage{placeins}
\usepackage{listings}

\lstdefinestyle{appendixprompt}{
  basicstyle=\ttfamily\small,
  columns=fullflexible,
  keepspaces=true,
  breaklines=true,
  breakatwhitespace=true,
  breakindent=0pt,
  showstringspaces=false,
  frame=single,
  framerule=0.3pt,
  framesep=3pt,
  xleftmargin=0pt,
  xrightmargin=0pt,
  aboveskip=4pt,
  belowskip=6pt,
  literate={—}{{\textemdash}}1
}

\usepackage{microtype}

\definecolor{cvprblue}{rgb}{0.21,0.49,0.74}
\usepackage[breaklinks,colorlinks,allcolors=cvprblue]{hyperref}

\def\confName{CVPR}
\def\confYear{2027}

\title{Next-Manip: On-Demand Agent--Policy Collaboration for Robotic Manipulation}

\author{
Jiayi Liu\textsuperscript{1}\thanks{Equal contribution.
\href{mailto:Jiayi.LIU@connect.hkust-gz.edu.cn}{\textcolor{black}{\nolinkurl{Jiayi.LIU@connect.hkust-gz.edu.cn}}};
\href{mailto:ke.ye@connect.hkust-gz.edu.cn}{\textcolor{black}{\nolinkurl{ke.ye@connect.hkust-gz.edu.cn}}}.}\quad
Ke Ye\textsuperscript{1}\footnotemark[1]\quad
Jiaming Zhou\textsuperscript{1}\quad
Yufei Huang\textsuperscript{2}\\
Haotian Liang\textsuperscript{2}\quad
Kun-Yu Lin\textsuperscript{3}\quad
Junwei Liang\textsuperscript{4}\thanks{Corresponding author.}\\
\textsuperscript{1}HKUST (Guangzhou)\quad
\textsuperscript{2}Tencent\quad
\textsuperscript{3}NTU\quad
\textsuperscript{4}HKUST
}

\hypersetup{
    pdftitle={Next-Manip: On-Demand Agent--Policy Collaboration for Robotic Manipulation},
    pdfauthor={Jiayi Liu, Ke Ye, Jiaming Zhou, Yufei Huang, Haotian Liang, Kun-Yu Lin, Junwei Liang}
}

\begin{document}
\maketitle

\begin{abstract}
General-purpose robotic manipulation requires task reasoning, failure recovery, and precise control, yet limited manipulation data make these abilities difficult to obtain from a single policy.
Recent multimodal agents show promise in task reasoning and recovery but lack reliable fine-grained control and incur high inference cost, whereas learned policies provide efficient fine-grained execution but struggle beyond their learned skills.
Although their complementary strengths suggest combining the two, doing so requires repeatedly deciding which executor should act as task demands change.
Instead of relying on the agent for each decision, we present Next-Manip, an on-demand agent--policy collaboration framework that decouples executor allocation from costly agent inference.
We propose an Agent Invocation Controller (AIC) that allocates control between the policy and the agent through a lightweight classifier of current execution demands.
We further introduce Task Dynamics Context (TDC), which packages informative manipulation changes from in-context examples into compact task references to support both agent reasoning and executor allocation.
Across two simulation benchmarks and real-world experiments, Next-Manip achieves state-of-the-art performance while reducing agent inference costs.
\end{abstract}
    
\section{Introduction}
\label{sec:intro}

\begin{figure}[t]
    \centering
    \includegraphics[width=\linewidth]{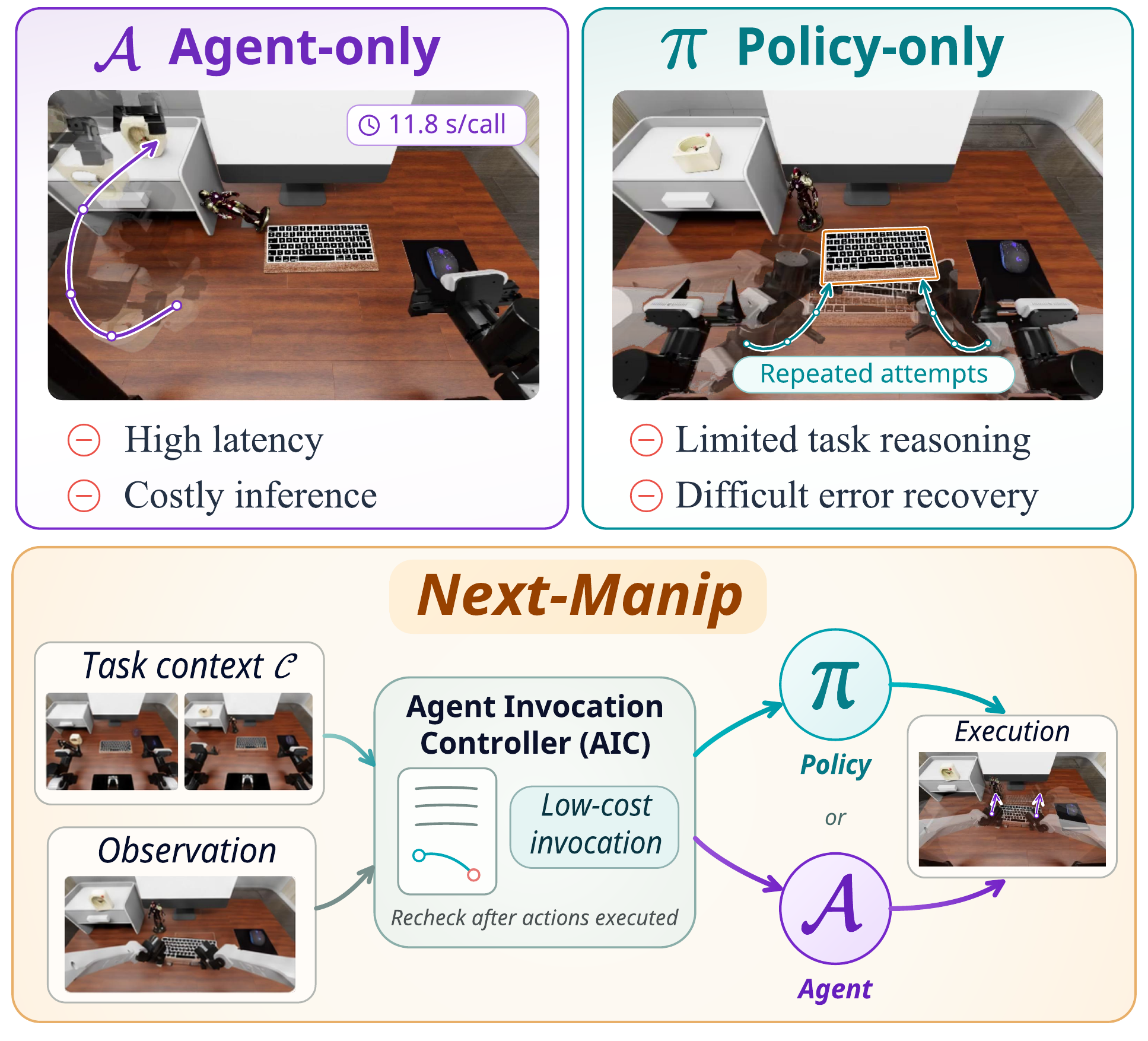}
    \captionsetup{skip=4pt}
    \caption{Next-Manip combines a reasoning agent and a manipulation policy through context-guided, on-demand collaboration. The agent invocation controller uses task context and observations to select an executor.}
    \label{fig:teaser}
    \vspace{-6pt}
\end{figure}

General-purpose robotic manipulation requires task understanding and reliable execution across diverse objects and environments~\cite{brohan2023rt,kim2024openvla,intelligence2025pi_,ye2026world}.
This calls for semantic and spatial reasoning, long-horizon planning, precise control, and error recovery~\cite{huang2023voxposer,goyal2023rvt,shi2025hi,zhao2023learning,chi2025diffusion,ma2026cyclevla}.
Learning-based approaches have made substantial progress in robotic manipulation~\cite{brohan2023rt,kim2024openvla,intelligence2025pi_,cheang2024gr}, supported by large-scale pretraining on diverse manipulation data~\cite{o2024open,contributors2026agibotworldcolosseolargescale,zhu2025unified}.
Despite these advances, integrating broad task competence and reliable control into a single policy still demands substantial further scaling of both manipulation data and general-purpose pretraining~\cite{intelligence2025pi_,contributors2026agibotworldcolosseolargescale,zhu2025unified}.
For example, long-horizon tasks and recovery from execution failures span a wide range of task stages and execution conditions~\cite{chen2026robodojo}, yet manipulation data covering these situations remain scarce and difficult to collect, limiting the breadth of experience available for learning general-purpose manipulation policies~\cite{intelligence2025pi_,black2024pi_0,khazatsky2024droid}.

Recent advances in multimodal reasoning agents, exemplified by GPT-6 Astra (hereafter Astra), offer a promising opportunity to bring general-purpose reasoning capabilities to robotic manipulation~\cite{zhang2026unexpected,sensenova2026gpt6astraachievedembodiedintelligence,chen2026show}.
Early manipulation studies show that these agents can interpret visual scenes, reason about task requirements, and use tools to execute robot actions~\cite{zhang2026unexpected,su2026astra,chen2026show}.
However, strong reasoning does not guarantee reliable fine-grained or dynamic control, and repeated agent calls also increase execution time and
cost~\cite{zhang2026unexpected,chen2026show,sensenova2026gpt6astraachievedembodiedintelligence,si2026fewertokensbetteraction}.
Learned manipulation policies, by contrast, can execute fine-grained actions at a low deployment cost~\cite{black2024pi_0,intelligence2025pi_,su2026astra}, but may struggle when task requirements exceed their learned manipulation skills~\cite{chen2026robodojo}.
These complementary strengths motivate combining reasoning agents with learned manipulation policies for effective and efficient task execution.

A central question we address is how to decide which executor should act as task demands change while keeping the overhead of collaboration low.
Recent work has explored using an agent to review and refine policy-generated actions~\cite{su2026astra}, demonstrating the potential of integrating task-level reasoning with policy execution.
However, the resulting hybrid does not consistently outperform the agent alone across tasks, suggesting that combining the two does not automatically yield more effective execution.
Moreover, every decision requires an agent call to review the policy's proposed actions, incurring repeated agent inference overhead even when no correction is needed.
These limitations motivate collaboration that adapts to the capabilities required by the task and its current execution state.
The agent should be able to lead execution when additional reasoning or recovery is needed, while the policy should proceed independently when it can handle the current demands.
We therefore seek an on-demand mechanism that allocates execution between the two without relying on costly agent reasoning for every invocation decision.

Both agentic reasoning and executor allocation require an understanding of task dynamics, including subgoal dependencies, operation preconditions, and expected effects.
In-context examples of the same task can provide concrete references for these aspects: they help the agent decompose tasks, interpret execution feedback, and reason about subsequent actions, while informing allocation decisions about which executor should act and when intervention is needed.
RoboICL demonstrates the value of multimodal demonstrations and interaction memory for embodied in-context learning~\cite{liu2026roboicl}.
However, retaining complete demonstrations increases the context burden, while simply sampling fewer frames may omit important changes in manipulation.
We therefore focus on organizing in-context examples around key changes in manipulation, preserving the information needed for both agent reasoning and executor allocation in a compact multimodal context.

To this end, we introduce \textbf{Next-Manip}, a unified framework for context-guided, on-demand collaboration between a general-purpose multimodal agent and a learned manipulation policy (\cref{fig:teaser}).
Specifically, we propose an \textbf{Agent Invocation Controller (AIC)} that separates inexpensive invocation decisions from costly agent reasoning to allocate control in response to changing task and execution demands.
We further introduce \textbf{Task Dynamics Context (TDC)}, which organizes task-relevant examples around key changes in manipulation into compact multimodal context to support both agent reasoning and executor allocation.
With both executors frozen, this design naturally extends to stronger agents and policies, enabling further performance gains.
Next-Manip thus brings heterogeneous capabilities into a unified collaboration framework, translating their complementary strengths into reliable and efficient task execution.

To summarize, our contributions are as follows:
\begin{itemize}
    \item We present Next-Manip, a unified framework combining general-purpose reasoning with learned manipulation capabilities through on-demand collaborative execution.
    \item We design AIC to dynamically select policy or agent execution through low-cost identification of evolving execution needs.
    \item We develop TDC to turn key manipulation changes in examples into concise task references for agent reasoning and allocation decisions.
    \item We evaluate Next-Manip on two simulation benchmarks and in real-world experiments, achieving the best overall performance among the compared methods while substantially reducing agent inference overhead.
\end{itemize}

\section{Related Work}
\label{sec:related_work}

\noindent\textbf{End-to-End Robotic Policies.}
End-to-end robotic policies learn to map observations and task conditions to actions from demonstrations.
Large-scale real-world datasets such as Open X-Embodiment and DROID~\cite{o2024open,khazatsky2024droid} provide diverse demonstrations, while RoboTwin 2.0~\cite{chen2025robotwin} expands training coverage through scalable simulation and domain randomization.
These resources support generalist policy learning across tasks and embodiments~\cite{team2024octo,liu2025rdt,zheng2026x}.
Using this experience for action supervision, vision-language-action models (VLA) such as RT-2, OpenVLA, $\pi_0$, and $\pi_{0.5}$ adapt pretrained vision-language representations to robot control~\cite{brohan2023rt,kim2024openvla,black2024pi_0,intelligence2025pi_,li2024cogact,shi2026memoryvla,yang2026eventvla}.
Alongside this use of vision-language priors, world-action models (WAM) learn world dynamics~\cite{zhou2026tau_0,cheang2024gr,bi2026motus,li2026causal,guo2026video,yuan2026fast,xu2026forcingworldactionmodeling,zhou2026zero} through joint video-action modeling~\cite{ye2026world} or adaptation of pretrained video models~\cite{kim2026cosmos}, connecting future prediction with action generation during training or inference.
Despite differences in model design, these policies learn manipulation skills from robot action data and can execute familiar tasks quickly with fine-grained control at low inference cost~\cite{shukor2025smolvla,kim2025fine}.

\noindent\textbf{Agentic Robotic Manipulation.}
Reasoning agents support robotic manipulation by interpreting tasks, planning actions, and responding to execution feedback~\cite{fu2026cap,chen2026embodied,yao2026profiling}.
Agents can delegate execution through language subgoals~\cite{hu2026matters,shi2025hi}, spatial guidance~\cite{li2025hamster}, multi-level commands~\cite{chen2026steerable}, or callable policies~\cite{lei2026towards,zhang2026harness}, and revise their plans from feedback~\cite{lei2026towards}.
When multiple policies are available, agents may select among them using prior evidence~\cite{chen2026roborouter} or coordinate their use across subtasks, as in RoboHarness~\cite{huang2026roboharness}.
Agents can also review policy-proposed actions~\cite{su2026astra,sensenova2026gpt6astraachievedembodiedintelligence} or generate commands directly through robot interfaces~\cite{zhang2026unexpected,chen2026agentic}.
Harnesses organize observations, tools, and feedback for direct control~\cite{guo2026transferring,chen2026show,qu2026roboharnesssimpleharnessoutperforms,yuan2026democratizing}, while predictive rehearsal helps refine agent-proposed actions~\cite{zhang2026world}.
These studies show that an agent can guide learned policy execution or assume direct control of the robot, but coordinating the two typically relies on agent inference.
Next-Manip instead uses a lightweight classifier to allocate control on demand between two direct executors, combining agent reasoning and corrective actions with fast, low-cost, fine-grained policy execution.

\noindent\textbf{In-Context Learning in Manipulation.}
In-context learning supplies task experience through demonstrations provided at inference time to guide robot action generation~\cite{li2026context,palma2026bimanual,bao2026recursive,fang2026world,chen2026zeva}.
Prior work either trains policies to condition on demonstrations~\cite{fu2024context} or encodes demonstration observations and actions as textual or visual prompts for general-purpose models~\cite{di2024keypoint,guo2026transferring}.
Making these examples useful involves both selecting relevant demonstrations and representing their contents compactly.
Retrieval draws examples from related tasks~\cite{zhou2026exploring}, while RoboPrompt retains keyframes around gripper changes and motion pauses~\cite{yin2025context}.
RoboICL~\cite{liu2026roboicl} further supplements demonstrations with interaction memory from the ongoing rollout.
In Next-Manip, demonstration context informs executor allocation as well as agent reasoning.
TDC constructs this shared reference from a single successful demonstration as a compact sequence of state--target pairs, keeping it small enough to accompany recurring allocation decisions.

\section{Next-Manip Framework}
\label{sec:method}

\begin{figure*}[t]
    \centering
    \includegraphics[width=\textwidth]{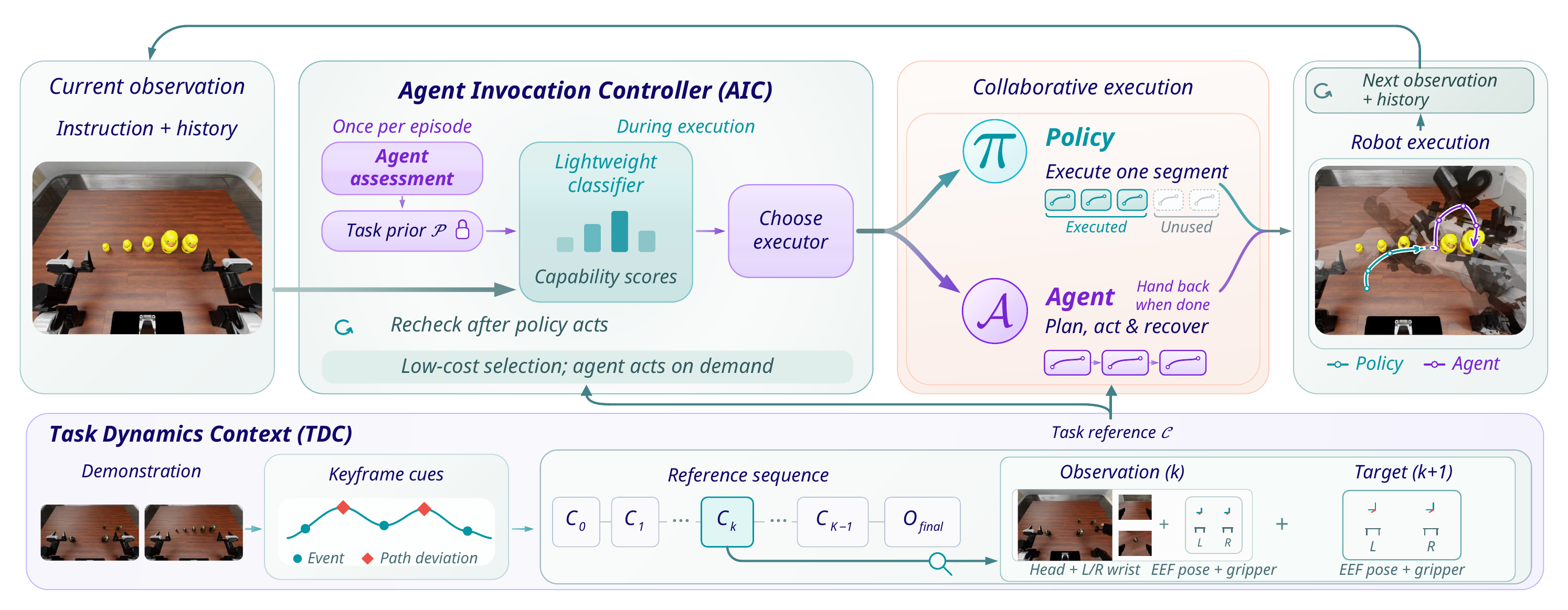}
    \captionsetup{skip=3pt}
    \caption{Overview of Next-Manip. Task Dynamics Context builds a task reference from a demonstration. The Agent Invocation Controller uses task information and observations to coordinate policy and agent execution.}
    \label{fig:method-pipeline}
\end{figure*}

Next-Manip is a unified framework that combines a general-purpose reasoning agent and a learned manipulation policy through context-guided, on-demand collaboration. Rather than committing either model to a fixed role, Next-Manip dynamically assigns control to the executor whose capabilities best match the current demands: the agent leads where task-level analysis or recovery is required, while the policy independently executes operations within its competence. Central to this design is the decoupling of \emph{who should act} (i.e., deciding which executor should lead) from \emph{how to act} (i.e., generating the robot actions), which shifts the frequent allocation decisions to a lightweight classifier and enables on-demand use of costly agent inference.

As illustrated in \cref{fig:method-pipeline}, the framework is realized by two complementary modules. The Task Dynamics Context (TDC) module organizes an in-context demonstration into a compact task reference that grounds both agent reasoning and executor allocation in the expected course of the task. Building on this reference, the Agent Invocation Controller (AIC) module employs a lightweight classifier to allocate control online and governs its handover as execution unfolds. In the following, we first formalize the collaborative execution setting (Sec.~\ref{sec:method:setup}), and then elaborate on the TDC module (Sec.~\ref{sec:method:reference}) and the AIC module (Sec.~\ref{sec:method:invocation}).

\subsection{Problem Formulation}
\label{sec:method:setup}

We consider robotic manipulation in which a robot follows a natural-language instruction $\ell$ through sequential interaction with its environment. At interaction round $t$, the robot receives an observation $o_t=(I_t,x_t)$, comprising multi-view RGB images $I_t$ and its proprioceptive state $x_t$, and advances the task by executing robot actions. The interaction history $\mathcal H_t$ records previous observations, executed actions, and control assignments within the current episode.

Next-Manip coordinates two frozen executors with complementary capabilities: a manipulation policy $\pi$ that provides efficient fine-grained control, and a reasoning agent $\mathcal A$ that provides general-purpose task reasoning and recovery. Each executor generates an action chunk $\mathbf A_t\in\mathbb R^{H_t\times d_{e_t}}$ in its native action representation, where $H_t$ is the predicted horizon and $d_{e_t}$ is the action dimension of the active executor $e_t\in\{\pi,\mathcal A\}$. Their collaboration is mediated by two modules. The Task Dynamics Context module $\mathcal M_{\mathrm{TDC}}$ converts a single successful in-context demonstration $\mathcal D$ of the task into a compact task reference $\mathcal C$, and the Agent Invocation Controller module $\mathcal M_{\mathrm{AIC}}$ determines which executor acts at each round and when control is handed over.

Given these definitions, the action chunk at round $t$ is generated by the active executor:
\begin{equation}
\label{eq:method:action}
\mathbf A_t=
\begin{cases}
\pi(\ell,o_t), & e_t=\pi,\\[2pt]
\mathcal A(\ell,o_t,\mathcal H_t,\mathcal C), & e_t=\mathcal A,
\end{cases}
\end{equation}
where the policy operates on its native inputs, while the agent additionally conditions on the interaction history and the task reference. The active executor is, in turn, determined by AIC from the instruction, the current observation, the history, and the task reference:
\begin{equation}
\label{eq:method:executor}
e_t=\mathcal M_{\mathrm{AIC}}(\ell,o_t,\mathcal H_t,\mathcal C),
\qquad e_t\in\{\pi,\mathcal A\}.
\end{equation}
Here, the task reference $\mathcal C$ is constructed by the TDC module from the in-context demonstration $\mathcal D$:
\begin{equation}
\label{eq:method:reference-construction}
\mathcal C=\mathcal M_{\mathrm{TDC}}(\mathcal D).
\end{equation}
After execution, the actions and the control assignment are appended to the history $\mathcal H_{t+1}$, and the process proceeds to the next round with observation $o_{t+1}$.

Under this formulation, the objective of Next-Manip is to assign control at each round to the executor whose capabilities match the current task demands, while keeping agent calls and execution latency low.

\subsection{Task Dynamics Context}
\label{sec:method:reference}

The TDC module is introduced to equip agent reasoning and executor allocation with procedural knowledge of the task, including the order of operations and their expected effects. The current observation describes the immediate state but does not explicitly specify how successive operations lead to task completion. We therefore draw on a single demonstration as a concrete reference for this progression. As a complete demonstration is too long to include as context, a straightforward way to reduce its size is to sample frames at uniform intervals. However, dense sampling repeatedly includes near-identical states, while sparse sampling risks omitting brief but important manipulation changes. TDC addresses this trade-off by organizing the reference around the \emph{dynamics} of the task: it concentrates on operational changes and abstracts the motion between consecutive keyframes into a single target.

\noindent\textbf{Dynamics-aware keyframe selection.}
TDC identifies keyframes from two complementary cues. The first is \emph{manipulation events}, defined as changes in the gripper command and the start or stop of end-effector motion. For each detected event, TDC retains the immediately preceding frame to preserve the conditions before the operational change. Event anchors alone, however, can miss pronounced path changes during continuous motion. The second cue is therefore \emph{path deviation}. Within intervals bounded by event anchors or demonstration endpoints, TDC additionally retains the maximum-deviation frame when the path deviates persistently from the straight segment joining the interval endpoints. TDC also retains the first and last demonstration frames as keyframes, marking the initial and completed task states.

\noindent\textbf{State--target abstraction.}
The selected keyframes partition the demonstration into intervals. To express the retained operations compactly, TDC replaces each interval's dense control record with a state--target pair consisting of its starting observation and a single target extracted from the next keyframe. Let $\tau_0<\cdots<\tau_K$ denote the selected keyframe indices. The pairs and the complete reference are defined as
\begin{equation}
\label{eq:method:task-reference}
\begin{aligned}
c_k &= \bigl(\tilde o_{\tau_k},\mathbf u_{\tau_{k+1}}\bigr),
\quad k=0,\ldots,K-1,\\
\mathcal C &= \bigl(c_0,\ldots,c_{K-1},\tilde o_{\tau_K}\bigr),
\end{aligned}
\end{equation}
where $\tilde o_{\tau_k}$ denotes the demonstration observation at keyframe $\tau_k$, comprising the images and robot state, and $\mathbf u_{\tau_{k+1}}$ specifies the absolute end-effector goal and gripper state extracted from the next keyframe. The resulting reference is a sparse, temporally ordered sequence describing the conditions from which operations begin and the targets they should reach, with the final observation specifying the completed state. Constructed once and fixed throughout the episode, the reference $\mathcal C$ provides task context for both agent reasoning and AIC's executor allocation.

\subsection{Agent Invocation Controller}
\label{sec:method:invocation}

The AIC module allocates control between the two complementary executors according to the capability that the current situation demands. Since allocation recurs throughout execution, entrusting it to the agent incurs costly inference at every decision even when additional agent reasoning is unnecessary. AIC therefore decouples recurring allocation decisions from agent reasoning: a lightweight classifier recognizes the capability currently required, and a handover mechanism determines when executor allocation is reconsidered.

\noindent\textbf{Lightweight capability classification.}
Executor allocation requires identifying the capability needed to advance the current operation, leaving the generation of robot actions to the selected executor. AIC expresses this allocation judgment as classification, using a lightweight, training-free classifier $g$ with prompts defining capability categories for executor allocation. The category set $\mathcal Y=\mathcal Y_\pi\cup\mathcal Y_{\mathcal A}$ reflects the executors' complementary strengths. Categories in $\mathcal Y_\pi$ describe operations within the policy's competence and are assigned to the policy $\pi$ to exploit its efficient learned control. Categories in $\mathcal Y_{\mathcal A}$ describe situations requiring additional semantic judgment, constraint planning, correction, or recovery, and are assigned to the agent $\mathcal A$. Abstracted prompt specifications are provided in Appendix~\ref{sec:appendix:prompts}.

Interpreting these capability requirements depends not only on the current state but also on the overall task context. To keep repeated classification queries compact, however, the classifier $g$ uses a limited recent interaction history $\bar{\mathcal H}_t$, comprising recent observations and execution records. AIC complements this local context with a task prior $\mathcal P$: before execution, the agent assesses the task's overall capability demands using the instruction $\ell$, the initial observation, and the TDC reference $\mathcal C$. This assessment remains fixed throughout the episode and conditions subsequent online judgments, allowing task-level guidance to be reused as execution unfolds. Whenever an executor is to be selected, the classifier returns
\begin{equation}
\label{eq:method:capability-classification}
(\hat y_t,p_t)
=
g\bigl(\ell,o_t,\bar{\mathcal H}_t,\mathcal C,\mathcal P\bigr),
\qquad \hat y_t\in\mathcal Y,
\end{equation}
where $p_t(y)$ is the probability assigned to category $y$ and $\hat y_t$ is the most probable category. AIC then assigns control to the corresponding executor, setting $e_t=\pi$ if $\hat y_t\in\mathcal Y_\pi$ and $e_t=\mathcal A$ otherwise.

\noindent\textbf{Asymmetric control handover.}
Once an executor is selected, AIC must also determine how long it retains control: long enough to make progress, yet not beyond the point at which the demands change. The executors differ in their ability to make this judgment, so AIC treats continuation and handover asymmetrically. The policy natively regenerates action chunks from updated observations, which provides natural points for re-examining the allocation. The agent, in contrast, may need several action chunks to complete a course of reasoning, correction, or recovery, and whether this course should continue depends on what it has already done. Its retained interaction history therefore supports making this continuation decision.

For the policy, the framework executes a bounded action segment before returning to the lightweight classifier, using the resulting observation $o_{t+1}$ to determine the next executor $e_{t+1}$.
When execution stalls across successive segments, control instead passes directly to the agent without a classifier query.
For the agent, continuation is decided autonomously alongside next-action planning, with the action chunk $\mathbf A_t$ and a handover indicator $b_t$ returned in a single response:
\begin{equation}
\label{eq:method:agent-handover}
(\mathbf A_t,b_t)
=\mathcal A(\ell,o_t,\mathcal H_t,\mathcal C,\mathcal P),
\qquad b_t\in\{0,1\},
\end{equation}
where $b_t=1$ hands control to the policy and $b_t=0$ retains agent control, completing Eq.~\eqref{eq:method:action} with the task prior $\mathcal P$. Issuing the handover decision jointly with the action plan requires no additional agent inference and avoids a separate classifier query. Upon handover, the policy executes a bounded segment before online executor allocation resumes. Together, TDC and AIC enable Next-Manip to draw on the executors' complementary strengths on demand, allowing either to lead as task requirements evolve while keeping coordination lightweight.

\section{Experiments}
\label{sec:experiments}

We evaluate whether on-demand agent--policy collaboration improves task execution while reducing reliance on costly agent reasoning, using RoboDojo~\cite{chen2026robodojo}, RMBench~\cite{chen2026rmbench}, and real-world manipulation tasks.

\subsection{Experimental Setup}
\label{sec:experiments:setup}
\label{sec:experiments:setup-benchmark}

\paragraph{Framework configuration.}
We use GPT-6 Astra as the reasoning agent and GPT-6 Luna through the Decisions API~\cite{openai2026decisions} as AIC's classifier. Next-Manip and all component variants use medium reasoning effort in simulation and real-world experiments, matching the Astra setting reported on the RoboDojo leaderboard~\cite{chen2026robodojo}. For consistency across executors, Astra and the policy execute the first ten actions per chunk. Reference demonstrations use layouts excluded from evaluation. These settings apply across all evaluations, with environment-specific configurations and protocols detailed below.

\begin{table*}[!t]
  \caption{RoboDojo~\cite{chen2026robodojo} success rate (\%) $\uparrow$ on 12 representative tasks. WALL: WALL-OSS-1.0-lite~\cite{zhai2025igniting}. $^*$ local, 5 trials/task; unmarked baselines: leaderboard. Ours: \textbf{best}, \underline{second-best}.}
  \label{tab:robodojo-sr}
  \centering
  \footnotesize
  \setlength{\tabcolsep}{1pt}
  \begin{tabularx}{\textwidth}{@{}ll*{12}{>{\centering\arraybackslash}X}@{\hspace{3pt}}c@{}}
    \toprule
    \textbf{\raisebox{-6pt}[0pt][0pt]{Category}} & \textbf{\raisebox{-6pt}[0pt][0pt]{Method}} & \multicolumn{3}{c}{\textbf{Generalization}} & \multicolumn{3}{c}{\textbf{Precision}} & \multicolumn{3}{c}{\textbf{Long-Horizon}} & \multicolumn{3}{c@{\hspace{3pt}}}{\textbf{Memory}} & \raisebox{-6pt}[0pt][0pt]{\textbf{Overall}} \\
    \cmidrule(lr){3-5}\cmidrule(lr){6-8}\cmidrule(lr){9-11}\cmidrule(lr){12-14}
      &  & Mugs & Pack & Largest & Charger & Tubes & Tower & Kong & Egg & Organize & Swap & Cover & \makebox[0pt]{Conveyor\hspace{3pt}} &  \\
    \midrule[0.07em]
    \raisebox{-7pt}[0pt][0pt]{VLA} & $\pi_{0.5}$~\cite{intelligence2025pi_} & 0.00 & 4.00 & 1.33 & 0.00 & 3.33 & 24.00 & 26.67 & 0.00 & 0.00 & 0.00 & 13.33 & 14.00 & 7.22 \\
      & Simate & 1.33 & 17.33 & 12.00 & 29.33 & 22.00 & 83.33 & 76.00 & 0.67 & 26.00 & 38.67 & 100.0 & 56.00 & 38.56 \\
      & WALL~\cite{zhai2025igniting} & 44.00 & 13.33 & 32.00 & 11.33 & 31.33 & 58.00 & 54.00 & 2.00 & 17.33 & 11.33 & 96.67 & 86.00 & 38.11 \\
    \midrule
    \raisebox{-4pt}[0pt][0pt]{WAM} & Awomo-0.5~\cite{team2026awomo} & 18.67 & 10.67 & 14.67 & 7.33 & 71.33 & 52.00 & 30.00 & 0.67 & 9.33 & 4.67 & 100.0 & 66.00 & 32.11 \\
      & VPP2~\cite{guo2026video} & 64.00 & 10.67 & 4.00 & 16.67 & 74.67 & 72.67 & 78.67 & 0.00 & 7.33 & 26.00 & 100.0 & 86.00 & 45.06 \\
    \midrule
    \raisebox{-7pt}[0pt][0pt]{Agent} & Astra & 0.00 & 0.00 & 68.00 & 0.00 & 0.00 & 2.00 & 0.00 & 0.00 & 0.00 & 0.00 & 46.00 & 62.00 & 14.83 \\
      & RoboICL$^*$~\cite{liu2026roboicl} & 0.00 & 20.00 & 60.00 & 40.00 & 40.00 & 100.0 & 0.00 & 0.00 & 20.00 & 0.00 & 100.0 & 100.0 & 40.00 \\
      & Q0-Preview & 28.00 & 12.00 & 88.00 & 44.00 & 46.00 & 72.00 & 42.00 & 0.00 & 36.00 & 90.00 & 100.0 & 84.00 & 53.50 \\
    \midrule
      & Astra+$\pi_{0.5}$$^*$~\cite{su2026astra} & 20.00 & 20.00 & 20.00 & 0.00 & 0.00 & 60.00 & 40.00 & 0.00 & 20.00 & 0.00 & 80.00 & 40.00 & 25.00 \\
      Hybrid & GO-3 & 0.00 & 0.00 & 76.00 & 70.00 & 92.00 & 100.0 & 100.0 & 0.00 & 16.00 & 36.00 & 100.0 & 28.00 & 51.50 \\
    \rowcolor{ProcessBlue!10}
    \cellcolor{white} & \textbf{Next-Manip} & 0.00 & \textbf{35.00} & \textbf{90.00} & \underline{50.00} & \underline{80.00} & \underline{95.00} & 45.00 & \textbf{10.00} & \underline{35.00} & \textbf{90.00} & \textbf{100.0} & 85.00 & \textbf{59.58} \\
    \bottomrule
  \end{tabularx}
\end{table*}

\paragraph{RoboDojo settings.}
RoboDojo~\cite{chen2026robodojo} is a benchmark for evaluating generalist robot manipulation policies. For a representative evaluation, we select three tasks from each of four capability categories on RoboDojo Standard: Generalization, Precision, Long-Horizon, and Memory. We choose the tasks in each category to cover low, middle, and high Astra performance, ranking all tasks by the Astra scores reported on the RoboDojo leaderboard~\cite{chen2026robodojo}, dividing them into three strata, and taking the median task from each. We adopt Simate, a strong and efficient VLA policy from the RoboDojo leaderboard, as the manipulation policy. Evaluation follows native success criteria and task-specific step limits. Each task is evaluated with 20 trials, two per layout (L0--L9). Success rate and native Score (0--100) are averaged equally across tasks. Locally reproduced baselines and ablation variants use five trials per task, one per layout (L0--L4).

\paragraph{RMBench settings.}
RMBench~\cite{chen2026rmbench} focuses on history-dependent manipulation in simulation. We evaluate all nine tasks, covering $M(1)$ and $M(n)$ memory requirements that involve one and multiple past observations, with a fixed EventVLA~\cite{yang2026eventvla} manipulation policy. Evaluation follows native success criteria and task-specific step limits. Next-Manip is evaluated with 20 trials per task, two runs for each of ten seeds; local Astra uses five trials per task. Task success rates are averaged equally.

\paragraph{Real-world settings.}
We conduct real-world experiments to examine whether the benefits of on-demand collaboration extend to physical execution. The platform is an xArm7 equipped with a DH Robotics PGI-140-80 gripper and front-view and wrist-mounted RGB cameras. We evaluate four tabletop tasks adapted from RoboDojo~\cite{chen2026robodojo}, \emph{Classify Objects}, \emph{Cover Blocks}, \emph{Stack Blocks By Language}, and \emph{General Pickup}, which cover category-based sorting, ordered manipulation after occlusion, and language-conditioned manipulation. The policy-only baseline is $\pi_{0.5}$~\cite{intelligence2025pi_}, fine-tuned for each task using LoRA~\cite{hu2021lora} on 50 demonstrations for 30,000 steps. Next-Manip adds GPT-6 Astra on top of the same checkpoint, with AIC allocating execution between the two and the TDC reference built from one of the same demonstrations. The two settings therefore differ only in whether collaboration is introduced, and each is evaluated with 20 trials per task. Task scenes, the experimental setup, and task details are provided in Appendix~\ref{sec:appendix:real-world}.

\begin{table*}[!tbp]
  \caption{RoboDojo~\cite{chen2026robodojo} score (0--100) $\uparrow$. Ours: \textbf{best}, \underline{second-best}.}
  \label{tab:robodojo-score}
  \centering
  \footnotesize
  \setlength{\tabcolsep}{1pt}
  \begin{tabularx}{\textwidth}{@{}ll*{12}{>{\centering\arraybackslash}X}@{\hspace{3pt}}c@{}}
    \toprule
    \textbf{\raisebox{-6pt}[0pt][0pt]{Category}} & \textbf{\raisebox{-6pt}[0pt][0pt]{Method}} & \multicolumn{3}{c}{\textbf{Generalization}} & \multicolumn{3}{c}{\textbf{Precision}} & \multicolumn{3}{c}{\textbf{Long-Horizon}} & \multicolumn{3}{c@{\hspace{3pt}}}{\textbf{Memory}} & \raisebox{-6pt}[0pt][0pt]{\textbf{Overall}} \\
    \cmidrule(lr){3-5}\cmidrule(lr){6-8}\cmidrule(lr){9-11}\cmidrule(lr){12-14}
      &  & Mugs & Pack & Largest & Charger & Tubes & Tower & Kong & Egg & Organize & Swap & Cover & \makebox[0pt]{Conveyor\hspace{3pt}} &  \\
    \midrule[0.07em]
    \raisebox{-7pt}[0pt][0pt]{VLA} & $\pi_{0.5}$~\cite{intelligence2025pi_} & 9.53 & 22.80 & 5.13 & 0.00 & 17.87 & 37.73 & 26.67 & 2.23 & 23.33 & 0.00 & 19.07 & 14.00 & 14.86 \\
      & Simate & 13.60 & 49.93 & 18.47 & 29.33 & 41.20 & 87.40 & 76.00 & 16.60 & 71.83 & 38.67 & 100.0 & 56.00 & 49.92 \\
      & WALL~\cite{zhai2025igniting} & 62.00 & 38.27 & 45.07 & 11.33 & 51.73 & 68.27 & 54.00 & 15.00 & 70.17 & 11.33 & 97.67 & 86.00 & 50.90 \\
    \midrule
    \raisebox{-4pt}[0pt][0pt]{WAM} & Awomo-0.5~\cite{team2026awomo} & 36.33 & 31.33 & 26.07 & 7.33 & 79.47 & 55.67 & 30.00 & 7.20 & 49.50 & 4.67 & 100.0 & 66.00 & 41.13 \\
      & VPP2~\cite{guo2026video} & 74.20 & 37.53 & 11.67 & 16.67 & 82.93 & 79.20 & 78.67 & 10.23 & 63.17 & 26.00 & 100.0 & 86.00 & 55.52 \\
    \midrule
    \raisebox{-7pt}[0pt][0pt]{Agent} & Astra & 0.60 & 11.00 & 76.00 & 0.00 & 6.00 & 16.40 & 0.00 & 5.50 & 39.50 & 0.00 & 49.30 & 62.00 & 22.19 \\
      & RoboICL$^*$~\cite{liu2026roboicl} & 6.00 & 42.00 & 74.00 & 40.00 & 60.00 & 100.0 & 0.00 & 20.00 & 70.00 & 0.00 & 100.0 & 100.0 & 51.00 \\
      & Q0-Preview & 49.80 & 34.20 & 89.40 & 44.00 & 64.40 & 77.00 & 42.00 & 8.20 & 79.00 & 90.00 & 100.0 & 84.00 & 63.50 \\
    \midrule
      & Astra+$\pi_{0.5}$$^*$~\cite{su2026astra} & 39.00 & 50.00 & 37.00 & 0.00 & 36.00 & 64.00 & 40.00 & 23.00 & 65.00 & 0.00 & 83.00 & 40.00 & 39.75 \\
      Hybrid & GO-3 & 3.00 & 8.60 & 82.60 & 70.00 & 95.20 & 100.0 & 100.0 & 5.40 & 74.00 & 36.00 & 100.0 & 28.00 & 58.57 \\
    \rowcolor{ProcessBlue!10}
    \cellcolor{white} & \textbf{Next-Manip} & 9.25 & \textbf{56.50} & \textbf{93.50} & \underline{50.00} & \underline{84.00} & \underline{95.50} & 45.00 & \textbf{27.00} & \underline{78.75} & \textbf{90.00} & \textbf{100.0} & 85.00 & \textbf{67.88} \\
    \bottomrule
  \end{tabularx}
\end{table*}

\subsection{Experimental Results}
\label{sec:experiments:results}
\label{sec:experiments:simulation}

\paragraph{Overall performance on RoboDojo.}
On RoboDojo~\cite{chen2026robodojo}, we compare Next-Manip with VLA policies, world-action models, and agent-based methods, including agent--policy hybrids (Tables~\ref{tab:robodojo-sr} and~\ref{tab:robodojo-score}). Unmarked baselines are published references from the RoboDojo leaderboard~\cite{chen2026robodojo}, while RoboICL~\cite{liu2026roboicl} and Astra + $\pi_{0.5}$~\cite{su2026astra} are reproduced locally from their official repositories under the original xhigh reasoning effort. Next-Manip achieves 59.58\% success and a Score of 67.88 across the twelve tasks, exceeding the published standalone results of Astra (14.83\%, 22.19) and Simate (38.56\%, 49.92). It further outperforms Q0-Preview, the top-ranked method on the RoboDojo leaderboard, which achieves 53.50\% success and a Score of 63.50. In summary, Next-Manip improves overall task execution by bringing agent reasoning and policy control into an on-demand collaboration framework.

\paragraph{Performance across executor strengths.}
To examine whether collaboration combines both executors' strengths, we compare performance on two task groups: those favoring Astra and those favoring Simate. In the Astra-favored group, Next-Manip achieves 87.50\% success, compared with Astra's 65.00\%; in the Simate-favored group, it achieves 54.00\%, compared with Simate's 39.47\%. Score follows the same trend, with gains in both task groups. In particular, \emph{Largest} and \emph{Tower} illustrate complementary executor strengths: Astra and Simate achieve published success rates of 68\% and 12\% on \emph{Largest}, versus 2\% and 83.33\% on \emph{Tower}~\cite{chen2026robodojo}; Next-Manip achieves 90\% and 95\%, respectively. Next-Manip thus combines the complementary strengths of the agent and policy to improve performance across tasks with different execution demands.
Additionally, we report executor allocation statistics in Appendix~\ref{sec:appendix:allocation} and qualitative analysis of control handover in Appendix~\ref{sec:appendix:qualitative}.

\begin{table*}[tbp]
  \caption{RMBench~\cite{chen2026rmbench} success rate (\%) $\uparrow$. $^*$ local, 5 trials/task. Ours: \textbf{best}, \underline{second-best}.}
  \label{tab:rmbench-sr}
  \centering
  \footnotesize
  \setlength{\tabcolsep}{2pt}
  \begin{tabularx}{\textwidth}{@{}ll*{9}{>{\centering\arraybackslash}X}@{\hspace{3pt}}c@{}}
    \toprule
    \textbf{\raisebox{-6pt}[0pt][0pt]{Category}} & \textbf{\raisebox{-6pt}[0pt][0pt]{Method}} & \multicolumn{5}{c}{\textbf{M(1)}} & \multicolumn{4}{c@{\hspace{3pt}}}{\textbf{M(n)}} & \raisebox{-6pt}[0pt][0pt]{\textbf{Overall}} \\
    \cmidrule(lr){3-7}\cmidrule(lr){8-11}
      &  & Pickup & Rearrange & PutBack & Swap-B & Swap-T & Battery & Ranking & Cover & Button &  \\
    \midrule[0.07em]
    \raisebox{-14pt}[0pt][0pt]{VLA} & $\pi_{0.5}$~\cite{intelligence2025pi_} & 9.0 & 13.0 & 11.0 & 24.0 & 15.0 & 16.0 & 6.0 & 0.0 & 0.0 & 10.4 \\
      & MemoryVLA~\cite{shi2026memoryvla} & 2.0 & 53.0 & 81.0 & 76.0 & 9.0 & 33.0 & 53.0 & 69.0 & 0.0 & 41.8 \\
      & Mem-0~\cite{chen2026rmbench} & 4.0 & 89.0 & 90.0 & 67.0 & 14.0 & 28.0 & 18.0 & 68.0 & 0.0 & 42.0 \\
      & EventVLA~\cite{yang2026eventvla} & 21.0 & 96.0 & 95.0 & 96.0 & 87.0 & 35.0 & 81.0 & 97.0 & 3.0 & 67.9 \\
    \midrule
    Agent & Astra$^*$ & 80.0 & 0.0 & 100.0 & 60.0 & 100.0 & 40.0 & 100.0 & 20.0 & 80.0 & 64.4 \\
    \midrule
    \rowcolor{ProcessBlue!10}
    \cellcolor{white}Hybrid & \textbf{Next-Manip} & \textbf{100.0} & \underline{95.0} & \textbf{100.0} & \underline{85.0} & \textbf{100.0} & \textbf{70.0} & \textbf{100.0} & \textbf{100.0} & \textbf{100.0} & \textbf{94.4} \\
    \bottomrule
  \end{tabularx}
\end{table*}

\paragraph{Results on RMBench.}
On RMBench~\cite{chen2026rmbench}, we compare against baselines compiled by EventVLA~\cite{yang2026eventvla} and a locally evaluated Astra. Overall, Next-Manip achieves 94.4\% success, compared with 67.9\% for EventVLA and 64.4\% for local Astra (Table~\ref{tab:rmbench-sr}). It achieves 100\% success on six of nine tasks, with improvements spanning both $M(1)$ and $M(n)$. At the task level, the two executors exhibit different strengths: on \emph{Button}, Astra and EventVLA achieve 80\% and 3\% success, respectively, while Next-Manip achieves 100\%; on \emph{Rearrange}, they achieve 0\% and 96\%, while Next-Manip achieves 95\%. Accordingly, Next-Manip extends the benefits of on-demand agent--policy collaboration across benchmarks and manipulation policies, supporting the broader applicability of the unified framework.

\paragraph{Real-world results.}
\label{sec:experiments:real}
Beyond simulation, we compare Next-Manip with policy-only $\pi_{0.5}$~\cite{intelligence2025pi_} on the four real-world tasks. Next-Manip achieves 90\%, 75\%, 80\%, and 65\% success on \emph{Classify Objects}, \emph{Cover Blocks}, \emph{Stack Blocks By Language}, and \emph{General Pickup}, respectively, compared with 45\%, 10\%, 40\%, and 20\% for policy-only $\pi_{0.5}$, raising the average success rate from 28.75\% to 77.50\%. These gains are obtained without changing the underlying manipulation policy weights. The improvements support on-demand collaboration as a way to complement a learned policy with general-purpose agent capabilities, without additional manipulation data or policy training.

\begin{figure}[!t]
  \centering
  \includegraphics[width=\linewidth]{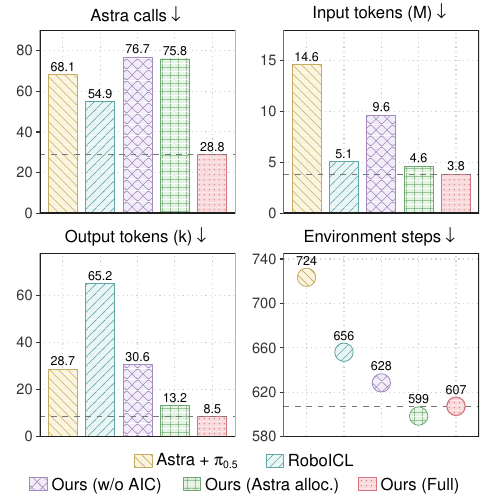}
  \vspace{-10pt}
  \captionsetup{skip=4pt}
  \caption{RoboDojo~\cite{chen2026robodojo} resource use per episode. Dashed: Full. Astra effort: Next-Manip medium; RoboICL~\cite{liu2026roboicl} and Astra + $\pi_{0.5}$~\cite{su2026astra} xhigh. Methods except Full use 5 trials/task.}
  \label{fig:robodojo-efficiency}
\end{figure}

\subsection{Efficiency and Execution Analysis}
\label{sec:experiments:efficiency}

\paragraph{Astra resource savings.}
The preceding results show the performance gains from collaboration, and we further assess its efficiency on RoboDojo~\cite{chen2026robodojo} in terms of model calls and token usage. Figure~\ref{fig:robodojo-efficiency} reports per-episode averages for Astra calls $\downarrow$ (lower is better), input tokens $\downarrow$, output tokens $\downarrow$, and environment steps $\downarrow$, including successful and failed trials. At medium reasoning effort, we make two comparisons. First, to measure the Astra resources saved by collaboration relative to standalone agent execution, we compare the full framework (Full) with w/o AIC, in which Astra executes alone with TDC. Full uses 28.80 Astra calls per episode, 62.5\% fewer than the 76.72 of w/o AIC, and recorded input and output tokens decrease by 60.1\% and 72.2\%, respectively. Second, to examine whether replacing Astra with the lightweight classifier for executor allocation sacrifices performance, we let Astra allocate executors instead (Astra alloc.). Its success rate remains comparable to that of Full (61.67\% vs.\ 59.58\%, Table~\ref{tab:robodojo-ablation}), while its Astra calls rise by 163.3\% to 75.83 per episode and its recorded input and output tokens increase by 20.7\% and 55.0\%, respectively. Taken together, these results indicate that on-demand coordination, which decouples executor allocation from Astra inference, improves task performance while reducing reliance on continued Astra execution.

\paragraph{Comparison with other methods.}
We further compare resource use and task performance with locally reproduced RoboICL~\cite{liu2026roboicl} and Astra + $\pi_{0.5}$~\cite{su2026astra}. At their original xhigh reasoning effort, RoboICL and Astra + $\pi_{0.5}$ use 54.88 and 68.07 Astra calls per episode and achieve 40.00\% and 25.00\% success, respectively. By comparison, Full achieves 59.58\% success at medium effort with 28.80 calls per episode. Its recorded input and output token usage is also lower than that of either method. In this comparison, resource savings accompany better task completion, supporting a more favorable balance between task performance and Astra resource demands with on-demand collaboration.

\paragraph{Execution and coordination overhead.}
Model-side savings must also be assessed alongside environment execution and coordination overhead. Full uses 607.29 environment steps per episode on average, close to w/o AIC's 628.38 and Astra alloc.'s 598.52, and below those of RoboICL~\cite{liu2026roboicl} and Astra + $\pi_{0.5}$~\cite{su2026astra}, showing that reduced Astra usage does not increase average environment steps. Beyond environment execution, AIC introduces additional model calls for online executor selection. These judgments are handled by the GPT-6 Luna classifier (Sec.~\ref{sec:method:invocation}), which makes 49.03 calls per episode on average with a recorded input of 0.342M tokens, or 8.9\% of Astra's input. Moreover, under official standard pricing, Luna's uncached input rate is 1/100 of Astra's, with no output-token charge for the Decisions API~\cite{openai2026pricing,openai2026decisions}, which keeps the classifier's input cost below 1\% of Astra's. By contrast, Astra alloc.\ requires more and costlier Astra input for the same judgments and responds more slowly, as the Decisions API is about 10 times faster than the Responses API used by Astra~\cite{openai2026decisions}. Frequent executor selection thus incurs only modest overhead, consistent with AIC's design of using lightweight judgments to support on-demand Astra execution.

\subsection{Ablations and Design Analysis}
\label{sec:experiments:analysis}

\paragraph{Ablation of TDC task references.}
Beyond efficiency, we further examine how each component contributes to task performance. To assess the role of task references, we compare Full with two variants (Table~\ref{tab:robodojo-ablation}), w/o TDC, which uses no reference, and Uniform reference, which replaces TDC with a reference sampled uniformly from the demonstration. Full achieves 59.58\% success and a Score of 67.88, higher than the 51.67\% and 61.58 of w/o TDC and the 55.00\% and 65.67 of Uniform reference, showing the benefit of TDC in collaborative execution. Standalone agent results provide complementary evidence: w/o AIC (Astra alone with TDC) achieves 43.33\% success and a Score of 51.33, exceeding the RoboDojo leaderboard Astra reference~\cite{chen2026robodojo} of 14.83\% and 22.19 at the same medium reasoning effort. It also performs comparably to locally reproduced RoboICL~\cite{liu2026roboicl} (40.00\%, 51.00), which uses xhigh reasoning effort. Together, these comparisons support the usefulness of task references for both agent execution and collaboration.
\begin{table}[!t]
  \caption{RoboDojo~\cite{chen2026robodojo} design analysis (12-task mean). Ref.: task reference; Alloc.: executor allocation; Unif.: uniform sampling; Med/Xh: medium/xhigh effort; SR: success rate; Score: native task score (0--100). $^*$ 5 trials/task; Astra only: leaderboard.}
  \label{tab:robodojo-ablation}
  \label{tab:robodojo-ablation-v2}
  \centering
  \footnotesize
  \setlength{\tabcolsep}{2.8pt}
  \begin{tabular}{@{}lcccccc@{}}
    \toprule
    \textbf{Configuration} & \textbf{Effort} & \textbf{Ref.} & \textbf{Policy} & \textbf{Alloc.} & \textbf{SR (\%)} & \textbf{Score} \\
    \midrule[0.07em]
    Astra only & Med & -- & -- & -- & 14.83 & 22.19 \\
    \midrule
    RoboICL$^*$~\cite{liu2026roboicl} & Xh & Unif. & -- & -- & 40.00 & 51.00 \\
    \midrule
    Next-Manip & & & & & & \\
    \addlinespace[2pt]
    \hspace{1em}\textit{w/o TDC}$^*$ & Med & -- & \checkmark & Luna & 51.67 & 61.58 \\
    \hspace{1em}\textit{Uniform reference}$^*$ & Med & Unif. & \checkmark & Luna & 55.00 & 65.67 \\
    \cmidrule(l{1em}r){1-7}
    \hspace{1em}\textit{w/o AIC}$^*$ & Med & TDC & -- & -- & 43.33 & 51.33 \\
    \hspace{1em}\textit{Random allocation}$^*$ & Med & TDC & \checkmark & Random & 30.00 & 39.92 \\
    \hspace{1em}\textit{Astra allocation}$^*$ & Med & TDC & \checkmark & Astra & 61.67 & 70.25 \\
    \cmidrule(l{1em}r){1-7}
    \rowcolor{ProcessBlue!10}
    \hspace{1em}\textbf{Full} & Med & TDC & \checkmark & Luna & 59.58 & 67.88 \\
    \bottomrule
  \end{tabular}
\end{table}

\paragraph{Ablation of AIC coordination.}
To assess the role of AIC's executor allocation, we retain TDC and compare Full with three variants (Table~\ref{tab:robodojo-ablation}): w/o AIC, in which Astra executes alone; Random alloc., which assigns each execution chunk to Astra or the policy with equal probability; and Astra alloc., in which Astra makes the allocation judgments. Full achieves 59.58\% success and a Score of 67.88, higher than both w/o AIC (43.33\%, 51.33) and Random alloc.\ (30.00\%, 39.92). Random alloc.\ even falls below w/o AIC, indicating that the gain arises from on-demand allocation rather than merely from an additional executor. Compared with Astra alloc.\ (61.67\%, 70.25), Full performs comparably while reducing reliance on costly, high-latency Astra inference (Sec.~\ref{sec:experiments:efficiency}). These ablations further clarify the role of each module, showing that collaboration benefits from on-demand allocation, which a lightweight classifier can provide without costly agent inference.

\section{Conclusion}
\label{sec:conclusion}

We present Next-Manip, an on-demand agent--policy collaboration framework that decouples executor allocation from costly agent inference.
Compact references distilled from demonstrations provide task context.
Lightweight online decisions allocate control between the policy and the agent as task demands change.
Experiments in simulation and on real robots demonstrate improved overall task performance, while efficiency analysis shows reduced reliance on costly agent inference.
Currently, Next-Manip adopts frozen models as executors, and its performance is therefore constrained by their capabilities.
Nevertheless, the framework can incorporate diverse agents and policies as they advance, supporting stronger collaborative execution.
As an early exploration of agent--policy collaboration, Next-Manip unifies heterogeneous reasoning and control capabilities at the execution level, offering a path toward manipulation systems that turn complementary model strengths into broader task competence.

{
    \small
    \bibliographystyle{arxivrefs}
    \bibliography{main}

@article{zhai2025igniting,
  title   = {Igniting {VLMs} Toward the Embodied Space},
  author  = {Zhai, Andy and Liu, Brae and Fang, Bruno and Cai, Chalse and Ma, Ellie and Yin, Ethan and Wang, Hao and Zhou, Hugo and Wang, James and Shi, Lights and Liang, Lucy and Wang, Make and Wang, Qian and Gan, Roy and Yu, Ryan and Li, Shalfun and Liu, Starrick and Chen, Sylas and Chen, Vincent and Xu, Zach},
  journal = {arXiv preprint arXiv:2509.11766},
  year    = {2025}
}

@misc{sensenova2026gpt6astraachievedembodiedintelligence,
      title={Has {GPT-6-Astra} Achieved Embodied Intelligence?},
      author={{SenseNova} and {Joint Research Team} and Zhongang Cai and Ziang Cao and Tailai Chen and Han Chong Chua and Linjun Dai and Hanming Deng and Hokin Deng and Haiwen Diao and Chenjing Ding and Weichen Fan and Xiao Fu and Zerui Gong and Chenyang Gu and Zexin He and Zhi Hou and Bo Li and Qinyue Li and Quanzhou Li and Xiaowan Li and Zhichao Liao and Dahua Lin and Zilong Lin and Bo Liu and Chunxiao Liu and Pengfei Liu and Tianqi Liu and Ziwei Liu and Chen Change Loy and Lewei Lu and Jiapeng Luo and Yuexiao Ma and Liang Pan and Xingang Pan and Oscar Qian and Ziyi Shang and Yuyao Shao and Dawei Su and Dacheng Tao and Qingyi Tao and Quan Wang and Ruisi Wang and Xiaogang Wang and Ziyi Wang and Beichen Wen and Donghao Wu and Jiaxu Wu and Silei Wu and Xiaoming Wu and Zhonghua Wu and Hongcan Xiao and Weiye Xu and Atsushi Yamashita and Lei Yang and Zhe Yang and Baiqiao Yin and Wanqi Yin and Hanfei Zhao and Tianbo Zhong and Long Zhuo and Jing Zuo},
      year={2026},
      eprint={2610.22471},
      howpublished={arXiv preprint arXiv:2610.22471},
      archivePrefix={arXiv},
      primaryClass={cs.RO},
      url={https://arxiv.org/abs/2610.22471},
}

@misc{xu2026forcingworldactionmodeling,
      title={{Next Forcing}: World Action Modeling with Multi-Chunk Prediction},
      author={Gangwei Xu and Qihang Zhang and Jiaming Zhou and Xing Zhu and Yujun Shen and Xin Yang and Yinghao Xu},
      year={2026},
      eprint={2606.11187},
      howpublished={arXiv preprint arXiv:2606.11187},
      archivePrefix={arXiv},
      primaryClass={cs.CV},
      url={https://arxiv.org/abs/2606.11187},
}

@misc{contributors2026agibotworldcolosseolargescale,
      title={{AgiBot World Colosseo}: A Large-scale Manipulation Platform for Scalable and Intelligent Embodied Systems},
      author={{AgiBot World Contributors} and Qingwen Bu and Jisong Cai and Li Chen and Xiuqi Cui and Yan Ding and Siyuan Feng and Shenyuan Gao and Xindong He and Xuan Hu and Xu Huang and Shu Jiang and Yuxin Jiang and Cheng Jing and Hongyang Li and Jialu Li and Chiming Liu and Yi Liu and Yuxiang Lu and Jianlan Luo and Ping Luo and Yao Mu and Yuehan Niu and Yixuan Pan and Jiangmiao Pang and Yu Qiao and Guanghui Ren and Cheng Ruan and Jiaqi Shan and Yongjian Shen and Chengshi Shi and Mingkang Shi and Modi Shi and Chonghao Sima and Jianheng Song and Huijie Wang and Wenhao Wang and Dafeng Wei and Chengen Xie and Guo Xu and Junchi Yan and Cunbiao Yang and Lei Yang and Shukai Yang and Maoqing Yao and Jia Zeng and Chi Zhang and Qinglin Zhang and Bin Zhao and Chengyue Zhao and Jiaqi Zhao and Jianchao Zhu},
      year={2026},
      eprint={2503.06669},
      howpublished={arXiv preprint arXiv:2503.06669},
      archivePrefix={arXiv},
      primaryClass={cs.RO},
      url={https://arxiv.org/abs/2503.06669},
}

@misc{qu2026roboharnesssimpleharnessoutperforms,
      title={{RoboHarness}: A Simple Harness Outperforms {VLA} and World Action Models},
      author={Bince Qu and Wei Chen and Bo Zhang},
      year={2026},
      eprint={2610.14018},
      howpublished={arXiv preprint arXiv:2610.14018},
      archivePrefix={arXiv},
      primaryClass={cs.RO},
      url={https://arxiv.org/abs/2610.14018},
}

@misc{si2026fewertokensbetteraction,
      title={Fewer Tokens, Better Action: {GPT-6 Astra} Robot Agents with 14\% Higher Success Rate but 65\% Fewer Tokens},
      author={Ruiyang Si and Jianxin Bi and Shunyu Yang and Rui Ni and Wenbo Huang and Qiang Wang and Shulong Jiang and Yintianrun Zhang and Feixiang Ruan and Kewei Zhang and Duomin Wang and Xiuyu Li and Haiwen Feng and Zhen Dong and Daquan Zhou},
      year={2026},
      eprint={2610.01939},
      howpublished={arXiv preprint arXiv:2610.01939},
      archivePrefix={arXiv},
      primaryClass={cs.CV},
      url={https://arxiv.org/abs/2610.01939},
}

@article{huang2023voxposer,
  title={{VoxPoser}: Composable {3D} value maps for robotic manipulation with language models},
  author={Huang, Wenlong and Wang, Chen and Zhang, Ruohan and Li, Yunzhu and Wu, Jiajun and Fei-Fei, Li},
  journal={arXiv preprint arXiv:2307.05973},
  year={2023}
}

@inproceedings{goyal2023rvt,
  title={{RVT}: Robotic view transformer for {3D} object manipulation},
  author={Goyal, Ankit and Xu, Jie and Guo, Yijie and Blukis, Valts and Chao, Yu-Wei and Fox, Dieter},
  booktitle={Conference on Robot Learning},
  pages={694--710},
  year={2023},
  organization={PMLR}
}

@article{zhao2023learning,
  title={Learning fine-grained bimanual manipulation with low-cost hardware},
  author={Zhao, Tony Z and Kumar, Vikash and Levine, Sergey and Finn, Chelsea},
  journal={arXiv preprint arXiv:2304.13705},
  year={2023}
}

@article{ma2026cyclevla,
  title={{CycleVLA}: Proactive self-correcting vision-language-action models via subtask backtracking and minimum {Bayes} risk decoding},
  author={Ma, Chenyang and Lu, Kai and Yang, Guangyu and Liu, Jiuming and Xu, Shitong and Byrne, Bill and Havoutis, Ioannis and Trigoni, Niki and Markham, Andrew},
  journal={arXiv preprint arXiv:2601.02295},
  year={2026}
}

@article{zhu2025unified,
  title={Unified world models: Coupling video and action diffusion for pretraining on large robotic datasets},
  author={Zhu, Chuning and Yu, Raymond and Feng, Siyuan and Burchfiel, Benjamin and Shah, Paarth and Gupta, Abhishek},
  journal={arXiv preprint arXiv:2504.02792},
  year={2025}
}

@inproceedings{o2024open,
  title={{Open X-Embodiment}: Robotic learning datasets and {RT-X} models: {Open X-Embodiment} collaboration 0},
  author={O’Neill, Abby and Rehman, Abdul and Maddukuri, Abhiram and Gupta, Abhishek and Padalkar, Abhishek and Lee, Abraham and Pooley, Acorn and Gupta, Agrim and Mandlekar, Ajay and Jain, Ajinkya and others},
  booktitle={2024 IEEE International Conference on Robotics and Automation (ICRA)},
  pages={6892--6903},
  year={2024},
  organization={IEEE}
}

@article{khazatsky2024droid,
  title={{DROID}: A large-scale in-the-wild robot manipulation dataset},
  author={Khazatsky, Alexander and Pertsch, Karl and Nair, Suraj and Balakrishna, Ashwin and Dasari, Sudeep and Karamcheti, Siddharth and Nasiriany, Soroush and Srirama, Mohan Kumar and Chen, Lawrence Yunliang and Ellis, Kirsty and others},
  journal={arXiv preprint arXiv:2403.12945},
  year={2024}
}

@article{team2024octo,
  title={Octo: An open-source generalist robot policy},
  author={Team, Octo Model and Ghosh, Dibya and Walke, Homer and Pertsch, Karl and Black, Kevin and Mees, Oier and Dasari, Sudeep and Hejna, Joey and Kreiman, Tobias and Xu, Charles and others},
  journal={arXiv preprint arXiv:2405.12213},
  year={2024}
}

@article{li2024cogact,
  title={{CogACT}: A foundational vision-language-action model for synergizing cognition and action in robotic manipulation},
  author={Li, Qixiu and Liang, Yaobo and Wang, Zeyu and Luo, Lin and Chen, Xi and Liao, Mozheng and Wei, Fangyun and Deng, Yu and Xu, Sicheng and Zhang, Yizhong and others},
  journal={arXiv preprint arXiv:2411.19650},
  year={2024}
}

@inproceedings{liu2025rdt,
  title={{RDT-1B}: a diffusion foundation model for bimanual manipulation},
  author={Liu, Songming and Wu, Lingxuan and Li, Bangguo and Tan, Hengkai and Chen, Huayu and Wang, Zhengyi and Xu, Ke and Su, Hang and Zhu, Jun},
  booktitle={International Conference on Learning Representations},
  volume={2025},
  pages={29982--30009},
  year={2025}
}

@article{shukor2025smolvla,
  title={{SmolVLA}: A vision-language-action model for affordable and efficient robotics},
  author={Shukor, Mustafa and Aubakirova, Dana and Capuano, Francesco and Kooijmans, Pepijn and Palma, Steven and Zouitine, Adil and Aractingi, Michel and Pascal, Caroline and Russi, Martino and Marafioti, Andres and others},
  journal={arXiv preprint arXiv:2506.01844},
  year={2025}
}

@article{kim2025fine,
  title={Fine-tuning vision-language-action models: Optimizing speed and success},
  author={Kim, Moo Jin and Finn, Chelsea and Liang, Percy},
  journal={arXiv preprint arXiv:2502.19645},
  year={2025}
}

@article{cheang2024gr,
  title={{GR-2}: A generative video-language-action model with web-scale knowledge for robot manipulation},
  author={Cheang, Chi-Lam and Chen, Guangzeng and Jing, Ya and Kong, Tao and Li, Hang and Li, Yifeng and Liu, Yuxiao and Wu, Hongtao and Xu, Jiafeng and Yang, Yichu and others},
  journal={arXiv preprint arXiv:2410.06158},
  year={2024}
}

@inproceedings{bi2026motus,
  title={Motus: A unified latent action world model},
  author={Bi, Hongzhe and Tan, Hengkai and Xie, Shenghao and Wang, Zeyuan and Huang, Shuhe and Liu, Haitian and Zhao, Ruowen and Feng, Yao and Xiang, Chendong and Rong, Yinze and others},
  booktitle={Proceedings of the IEEE/CVF Conference on Computer Vision and Pattern Recognition},
  pages={35101--35113},
  year={2026}
}

@article{ye2026world,
  title={World action models are zero-shot policies},
  author={Ye, Seonghyeon and Ge, Yunhao and Zheng, Kaiyuan and Gao, Shenyuan and Yu, Sihyun and Kurian, George and Indupuru, Suneel and Tan, You Liang and Zhu, Chuning and Xiang, Jiannan and others},
  journal={arXiv preprint arXiv:2602.15922},
  year={2026}
}

@article{li2026causal,
  title={Causal world modeling for robot control},
  author={Li, Lin and Zhang, Qihang and Luo, Yiming and Yang, Shuai and Wang, Ruilin and Han, Fei and Yu, Mingrui and Gao, Zelin and Xue, Nan and Zhu, Xing and others},
  journal={arXiv preprint arXiv:2601.21998},
  year={2026}
}

@article{kim2026cosmos,
  title={{Cosmos Policy}: Fine-tuning video models for visuomotor control and planning},
  author={Kim, Moo Jin and Gao, Yihuai and Lin, Tsung-Yi and Lin, Yen-Chen and Ge, Yunhao and Lam, Grace and Liang, Percy and Song, Shuran and Liu, Ming-Yu and Finn, Chelsea and others},
  journal={arXiv preprint arXiv:2601.16163},
  year={2026}
}

@article{yuan2026fast,
  title={{Fast-WAM}: Do world action models need test-time future imagination?},
  author={Yuan, Tianyuan and Dong, Zibin and Liu, Yicheng and Zhao, Hang},
  journal={arXiv preprint arXiv:2603.16666},
  year={2026}
}

@article{chen2025robotwin,
  title={{RoboTwin 2.0}: A scalable data generator and benchmark with strong domain randomization for robust bimanual robotic manipulation},
  author={Chen, Tianxing and Chen, Zanxin and Chen, Baijun and Cai, Zijian and Liu, Yibin and Li, Zixuan and Liang, Qiwei and Lin, Xianliang and Ge, Yiheng and Gu, Zhenyu and others},
  journal={arXiv preprint arXiv:2506.18088},
  year={2025}
}

@article{chen2026show,
  title={{Show-Harness}: Just a {VLM} agent can play robots},
  author={Chen, Yanzhe and Bai, Zechen and Cao, Zhijun and Zeng, Wenzheng and Lin, Kevin Qinghong and Lin, Yiqi and Liang, Guoqiang and Ma, Kevin Yuchen and Huang, Qiming and Shou, Mike Zheng},
  journal={arXiv preprint arXiv:2609.10522},
  year={2026}
}

@article{huang2026roboharness,
  title={{RoboHarness}: Memory-driven orchestration of heterogeneous robot policies for long-horizon planning},
  author={Huang, Jinbang and Li, Zhiyuan and Hu, Yuanzhao and Qi, Ran and Xiao, Yixin and Coates, Mark and Cao, Tongtong and Zhang, Zhanguang and Zhang, Yingxue},
  journal={arXiv preprint arXiv:2607.18060},
  year={2026}
}

@article{zhang2026world,
  title={{World Action Agent}: Harnessing {VLMs} for Robot Manipulation via World Action Rehearsal},
  author={Zhang, Yehang and Huang, Haojian and Chang, Yifan and Su, Jianchong and Zhou, Bohan and Xu, Yingjie and Chen, Wosong and Zhou, Tianhao and Wang, Chenxu and Zhang, Tianyi and others},
  journal={arXiv preprint arXiv:2609.29964},
  year={2026}
}

@article{fu2026cap,
  title={{CaP-X}: A framework for benchmarking and improving coding agents for robot manipulation},
  author={Fu, Letian and Yu, Justin and El-Refai, Karim and Kou, Ethan and Xue, Haoru and Huang, Huang and Xiao, Wenli and Wang, Guanzhi and Niu, Dantong and Li, Fei-Fei and others},
  journal={arXiv preprint arXiv:2603.22435},
  year={2026}
}

@article{zhang2026harness,
  title={{Harness VLA}: Steering frozen {VLAs} into reliable manipulation primitives via memory-guided agents},
  author={Zhang, Yixian and Zhang, Huanming and Gao, Feng and Li, Xiao and Liu, Zhihao and Nie, Yi and Zhu, Chunyang and Qiu, Jiaxing and Yan, Yuchen and Liu, Jiyuan and others},
  journal={arXiv preprint arXiv:2607.08448},
  year={2026}
}

@article{lei2026towards,
  title={Towards Long-horizon Embodied Agents with Tool-Aligned Vision-Language-Action Models},
  author={Lei, Zixing and Liu, Changxing and Xiong, Yichen and Xiong, Minhao and Ding, Yuanzhuo and Zhang, Zhipeng and Li, Weixin and Chen, Siheng},
  journal={arXiv preprint arXiv:2605.13119},
  year={2026}
}

@article{chen2026steerable,
  title={Steerable vision-language-action policies for embodied reasoning and hierarchical control},
  author={Chen, William and Bhatia, Jagdeep Singh and Glossop, Catherine and Mathihalli, Nikhil and Doshi, Ria and Tang, Andy and Driess, Danny and Pertsch, Karl and Levine, Sergey},
  journal={arXiv preprint arXiv:2602.13193},
  year={2026}
}

@article{chen2026roborouter,
  title={{RoboRouter}: Training-free policy routing for robotic manipulation},
  author={Chen, Yiteng and Cao, Zhe and Ren, Hongjia and Yang, Chenjie and Li, Wenbo and Wang, Shiyi and Wang, Yemin and Zhang, Li and Shao, Yanming and Zhao, Zhenjun and others},
  journal={arXiv preprint arXiv:2603.07892},
  year={2026}
}

@article{hu2026matters,
  title={What Matters in Orchestrating Robot Policies: A Systematic Study of Hierarchical {VLA} Agents},
  author={Hu, Jiaheng and Shridhar, Mohit and Lu, Caden and Shah, Dhruv and Chiang, Hao-Tien Lewis and Tan, Jie and Xie, Annie},
  journal={arXiv preprint arXiv:2606.10267},
  year={2026}
}

@article{chen2026agentic,
  title={On Agentic Manipulation Policies},
  author={Chen, Chen and Bagherieh, Omid and Balaji, Naveen Nikamanth Appiah and Gonzalez, Jesse T and Heravi, Negin and Islam, Fahad and Li, Xueting and Regli, Jean-Baptiste and Ren, Kejia and {\v{S}}vec, Petr and others},
  journal={arXiv preprint arXiv:2611.09892},
  year={2026}
}

@article{yuan2026democratizing,
  title={Democratizing Robot-Use Agents: A Unified Harness with Open Models},
  author={Yuan, Jiarui and Zuo, Yuxin and Huang, Xiaoyu and Zhang, Wenbo and Park, Kwanyoung and Yang, Junlin and Zhang, Tianxiang and Chen, Ruoqu and Ouyang, Yutao and He, Bingxiang and others},
  journal={arXiv preprint arXiv:2611.01560},
  year={2026}
}

@article{yao2026profiling,
  title={Profiling a frontier agent with {EBench}: From direct control to harness with policy},
  author={Yao, Yifei and Zou, Yude and Jiao, Guanbo and Gao, Xing and Gao, Ning and Wang, Yukai and Chen, Jiantong and Zhu, Zihou and Li, Xinyu and Yan, Ningyu and others},
  journal={arXiv preprint arXiv:2610.35832},
  year={2026}
}

@article{chen2026embodied,
  title={{Embodied Playground}: A Long-Horizon Agentic Robotics Benchmark for Physical, Strategic, and Spatial Reasoning},
  author={Chen, Feng and Zhou, Pei and He, Lixuan and Yao, Yunchao and Sun, Li and Gao, Shenghua and Ding, Mingyu and Zhang, Kaifeng and Zhou, Kaichen and Ma, Yi},
  journal={arXiv preprint arXiv:2610.27039},
  year={2026}
}

@article{bao2026recursive,
  title={Recursive Video In-Context Learning for Agentic Robot},
  author={Bao, Wenrui and Liu, Xinxin and Xu, Bingxin and Shang, Yuzhang},
  journal={arXiv preprint arXiv:2610.06843},
  year={2026}
}

@article{fang2026world,
  title={World In-Context Learning for Agentic Robot Manipulation},
  author={Fang, Heng and Guo, Zehui and Zhang, Yuehan and Du, Mingyang and Liang, Dingkang and Bai, Xiang},
  journal={arXiv preprint arXiv:2611.12885},
  year={2026}
}

@article{chen2026zeva,
  title={Zeva: In-context causal learning for generalizable embodied manipulation},
  author={Chen, Fu and Ding, Xin and Huang, Bingjia and Li, Xiangyu and Wang, Mingju and He, Jiawei and Li, Kun and Sun, Wei and Liu, Yunxin and Wu, Hao and others},
  journal={arXiv preprint arXiv:2608.30880},
  year={2026}
}

@article{palma2026bimanual,
  title={Bimanual Robot Manipulation via Multi-Agent In-Context Learning},
  author={Palma, Alessio and Spinelli, Indro and Prasad, Vignesh and Scofano, Luca and Jin, Yufeng and Chalvatzaki, Georgia and Galasso, Fabio},
  journal={arXiv preprint arXiv:2604.20348},
  year={2026}
}

@article{li2026context,
  title={In-context Robot Learning Made Simple: A Democratized Recipe for Manipulation Tasks},
  author={Li, Minxing and Han, Minghao and Zhao, Weizhi and Wang, Hanwen and Liu, Xiangshuo and Shang, Shuyao and Zhou, Jingxiang and Sun, Mingchao and Pan, Hongyu and Xu, Mu and others},
  journal={arXiv preprint arXiv:2609.38173},
  year={2026}
}

@inproceedings{shi2026memoryvla,
  title={{MemoryVLA}: Perceptual-cognitive memory in vision-language-action models for robotic manipulation},
  author={Shi, Hao and Xie, Bin and Liu, Yingfei and Sun, Lin and Liu, Fengrong and Wang, Tiancai and Zhou, Erjin and Fan, Haoqiang and Zhang, Xiangyu and Huang, Gao},
  booktitle={International Conference on Learning Representations},
  volume={2026},
  pages={18567--18602},
  year={2026}
}

@article{chen2026rmbench,
  title={{RMBench}: Memory-dependent robotic manipulation benchmark with insights into policy design},
  author={Chen, Tianxing and Wang, Yuran and Li, Mingleyang and Qin, Yan and Shi, Hao and Li, Zixuan and Hu, Yifan and Zhang, Yingsheng and Wang, Kaixuan and Chen, Yue and others},
  journal={arXiv preprint arXiv:2603.01229},
  year={2026}
}

@article{yang2026eventvla,
  title={{EventVLA}: Event-Driven Visual Evidence Memory for Long-Horizon Vision-Language-Action Policies},
  author={Yang, Ganlin and Tu, Zhangzheng and Yang, Yuqiang and Mao, Sitong and Dong, Junyi and Chen, Tianxing and Peng, Jiaqi and Xiong, Jing and Cao, Jiafei and Dai, Jifeng and others},
  journal={arXiv preprint arXiv:2606.20092},
  year={2026}
}

@article{guo2026video,
  title={{Video Prediction Policy 2}: Predict better, act better},
  author={Guo, Yanjiang and Yan, Haodong and Zhong, Zhide and Zhang, Zhongru and Yang, Qingyuan and Lu, Qingzhou and Chen, Xiaoyu and Wang, Yen-Jen and Deng, Shuying and Yang, Chenghan and others},
  journal={arXiv preprint arXiv:2610.10270},
  year={2026}
}

@inproceedings{zheng2026x,
  title={{X-VLA}: Soft-prompted transformer as scalable cross-embodiment vision-language-action model},
  author={Zheng, Jinliang and Li, Jianxiong and Wang, Zhihao and Liu, Dongxiu and Kang, Xirui and Feng, Yuchun and Zheng, Yinan and Zou, Jiayin and Chen, Yilun and Zeng, Jia and others},
  booktitle={International Conference on Learning Representations},
  volume={2026},
  pages={60580--60606},
  year={2026}
}

@article{chi2025diffusion,
  title={{Diffusion Policy}: Visuomotor policy learning via action diffusion},
  author={Chi, Cheng and Xu, Zhenjia and Feng, Siyuan and Cousineau, Eric and Du, Yilun and Burchfiel, Benjamin and Tedrake, Russ and Song, Shuran},
  journal={The International Journal of Robotics Research},
  volume={44},
  number={10-11},
  pages={1684--1704},
  year={2025},
  publisher={Sage Publications, Sage UK: London, England}
}

@article{kim2024openvla,
  title={{OpenVLA}: An open-source vision-language-action model},
  author={Kim, Moo Jin and Pertsch, Karl and Karamcheti, Siddharth and Xiao, Ted and Balakrishna, Ashwin and Nair, Suraj and Rafailov, Rafael and Foster, Ethan and Lam, Grace and Sanketi, Pannag and others},
  journal={arXiv preprint arXiv:2406.09246},
  year={2024}
}

@article{zhang2026unexpected,
  title={An unexpected robot policy: Early evaluations of {GPT-6 Astra} on {RoboDojo} and beyond},
  author={Zhang, Wenbo and Wang, Kaixuan and Ouyang, Yutao and Huang, Xiaoyu and Li, Liyang and Su, Kailun and Jin, Weiyang and Chai, Wenhao and Liang, Haotian and Dou, Zhiyang and others},
  journal={arXiv preprint arXiv:2609.24170},
  year={2026}
}

@article{chen2026robodojo,
  title={{RoboDojo}: A unified sim-and-real benchmark for comprehensive evaluation of generalist robot manipulation policies},
  author={Chen, Tianxing and Chen, Yue and Li, Zixuan and Tang, Junyuan and Su, Kailun and Lu, Haoran and Wan, Weijie and Chen, Baijun and Liu, Songling and Yan, Haowen and others},
  journal={arXiv preprint arXiv:2607.04434},
  year={2026}
}

@article{liu2026roboicl,
  title={{RoboICL}: Embodied in-context learning with {GPT-6 Astra}},
  author={Liu, Fangcheng and Shen, Yeqing and Cheng, Anda and Mi, Weishi and Tang, Chao and Liu, Chenyuan and Xiang, Yushun and Li, Tingguang and Li, Yong-Lu and Tang, Yehui},
  journal={arXiv preprint arXiv:2609.34261},
  year={2026}
}

@misc{su2026astra,
  title        = {{GPT 6 Astra} as an Embodied Policy},
  author       = {Su, Jiayi and Zheng, Yixin and Yan, Mi and Yi, Li and Zhang, Zhizheng and Wang, He},
  year         = {2026},
  howpublished = {Technical report and code},
  url          = {https://github.com/anonymous-report-421/eval-of-gpt-6-astra-as-policy}
}

@article{intelligence2025pi_,

  title={{$\pi_{0.5}$}: a Vision-Language-Action Model with Open-World Generalization},
  author={Intelligence, Physical and Black, Kevin and Brown, Noah and Darpinian, James and Dhabalia, Karan and Driess, Danny and Esmail, Adnan and Equi, Michael and Finn, Chelsea and Fusai, Niccolo and others},
  journal={arXiv preprint arXiv:2504.16054},
  year={2025}
}

@article{hu2021lora,
  title={{LoRA}: Low-rank adaptation of large language models},
  author={Hu, Edward J and Shen, Yelong and Wallis, Phillip and Allen-Zhu, Zeyuan and Li, Yuanzhi and Wang, Shean and Wang, Lu and Chen, Weizhu},
  journal={arXiv preprint arXiv:2106.09685},
  year={2021}
}

@article{black2024pi_0,

  title={{$\pi_0$}: A Vision-Language-Action Flow Model for General Robot Control},
  author={Black, Kevin and Brown, Noah and Driess, Danny and Esmail, Adnan and Equi, Michael and Finn, Chelsea and Fusai, Niccolo and Groom, Lachy and Hausman, Karol and Ichter, Brian and others},
  journal={arXiv preprint arXiv:2410.24164},
  year={2024}
}

@article{brohan2023rt,
  title={{RT-2}: Vision-language-action models transfer web knowledge to robotic control},
  author={Brohan, Anthony and Brown, Noah and Carbajal, Justice and Chebotar, Yevgen and Chen, Xi and Choromanski, Krzysztof and Ding, Tianli and Driess, Danny and Dubey, Avinava and Finn, Chelsea and others},
  journal={arXiv preprint arXiv:2307.15818},
  year={2023}
}

@article{fu2024context,
  title={In-context imitation learning via next-token prediction},
  author={Fu, Letian and Huang, Huang and Datta, Gaurav and Chen, Lawrence Yunliang and Panitch, William Chung-Ho and Liu, Fangchen and Li, Hui and Goldberg, Ken},
  journal={arXiv preprint arXiv:2408.15980},
  year={2024}
}

@article{di2024keypoint,
  title={Keypoint action tokens enable in-context imitation learning in robotics},
  author={Di Palo, Norman and Johns, Edward},
  journal={arXiv preprint arXiv:2403.19578},
  year={2024}
}

@inproceedings{yin2025context,
  title={In-context learning enables robot action prediction in {LLMs}},
  author={Yin, Yida and Wang, Zekai and Sharma, Yuvan and Niu, Dantong and Darrell, Trevor and Herzig, Roei},
  booktitle={2025 IEEE International Conference on Robotics and Automation (ICRA)},
  pages={8972--8979},
  year={2025},
  organization={IEEE}
}

@article{shi2025hi,
  title={{Hi Robot}: Open-ended instruction following with hierarchical vision-language-action models},
  author={Shi, Lucy Xiaoyang and Ichter, Brian and Equi, Michael and Ke, Liyiming and Pertsch, Karl and Vuong, Quan and Tanner, James and Walling, Anna and Wang, Haohuan and Fusai, Niccolo and others},
  journal={arXiv preprint arXiv:2502.19417},
  year={2025}
}

@inproceedings{li2025hamster,
  title={{HAMSTER}: Hierarchical action models for open-world robot manipulation},
  author={Li, Yi and Deng, Yuquan and Zhang, Jesse and Jang, Joel and Memmel, Marius and Garrett, Caelan and Ramos, Fabio and Fox, Dieter and Li, Anqi and Gupta, Abhishek and others},
  booktitle={International Conference on Learning Representations},
  volume={2025},
  pages={24040--24068},
  year={2025}
}

@article{zhou2026tau_0,

  title={{$\tau_0$-WM}: A Unified Video-Action World Model for Robotic Manipulation},
  author={Zhou, Pengfei and Chen, Shengcong and Chen, Di and Wang, Jiaxu and Jin, Rongjun and Zhu, Bingwen and Pan, Yike and Gu, Songen and Wang, Kuanning and Nan, Shufeng and others},
  journal={arXiv preprint arXiv:2606.01027},
  year={2026}
}

@article{zhou2026exploring,
  title={Exploring the limits of vision-language-action manipulation in cross-task generalization},
  author={Zhou, Jiaming and Ye, Ke and Liu, Jiayi and Ma, Teli and Wang, Zifan and Qiu, Ronghe and Lin, Kun-Yu and Zhao, Zhilin and Liang, Junwei},
  journal={Advances in Neural Information Processing Systems},
  volume={38},
  pages={139899--139927},
  year={2026}
}

@article{guo2026transferring,
  title={Transferring the intelligence of {VLMs} to robotic control},
  author={Guo, Meng-Hao and Mo, Zhe-Han and Wang, Jia-Jun and Zhang, Yi and Wang, Kejin and Deng, Yi-Xuan and Zhang, Jia-Peng and Rao, Yongming and Hu, Shi-Min},
  journal={arXiv preprint arXiv:2609.22966},
  year={2026}
}

@article{zhou2026zero,
  title={{Zero-WAM}: In-context world-action modeling from human videos for open-ended task generalization},
  author={Zhou, Jiaming and Zhang, Qihang and Xu, Gangwei and Fan, Cunxin and Zhao, Yujie and Wang, Ruilin and Luo, Yiming and Yang, Shuai and Zhu, Xing and Shen, Yujun and others},
  journal={arXiv preprint arXiv:2608.26103},
  year={2026}
}

@misc{team2026awomo,
  title={{Awomo-0.5} Technical Report},
  author={Team, Awomo-WAM},
  year={2026},
  howpublished={Technical report},
  note={\url{https://awomo-westlakedi.github.io/Awomo-0.5-page/assets/awomo-0.5-report.pdf}}
}

@misc{openai2026pricing,
  author = {{OpenAI}},
  title = {{OpenAI API Pricing}},
  year = {2026},
  howpublished = {\url{https://developers.openai.com/api/docs/pricing}},
  note = {Accessed November 10, 2026}
}

@misc{openai2026decisions,
  author = {{OpenAI}},
  title = {{Decisions API}},
  year = {2026},
  howpublished = {\url{https://developers.openai.com/api/docs/guides/decisions}},
  note = {Accessed November 10, 2026}
}
}

\clearpage
\appendix
\maketitlesupplementary
\section{Executor Allocation Prompts}
\label{sec:appendix:prompts}

The following abstracted prompt specifications summarize the capability classifier and task-prior assessment, omitting task-specific examples and interface formatting.

\subsection{Lightweight Capability Classification}
\label{sec:appendix:capability-prompts}

The classifier (GPT-6 Luna via the Decisions API) identifies current capability requirements, while AIC maps its output to an executor and handles control handover.

\begin{lstlisting}[style=appendixprompt]
ROLE
Identify the capability needed to advance the current task. Prioritize successful task execution, then efficiency. Select a capability category; leave action generation to the selected executor.

INPUTS
- Task instruction and public task conditions.
- Current RGB observations and robot state.
- Recent observations, execution records, and feedback.
- The fixed episode-level capability prior.
- The compact demonstration reference constructed by TDC.

CAPABILITY PRINCIPLES
1. Routine execution: prefer Policy when familiar manipulation skills support a clear continuation. Repetition, task duration, and fine alignment alone do not justify Astra. Preserve effective policy execution.
2. Reasoning: involve Astra when execution requires task-specific selection, assignment, ordering, calculation, counting, confirmation, or retained reference information beyond ordinary geometric grounding. Consider the coherent operation, not only the next motion. Routine-looking approach, contact, or release motions may remain part of an intervention that requires reasoning.
3. Physical process constraints: involve Astra when an explicit condition must hold during motion and requires planning about contact, support, orientation, or coordinated paths. Distinguish such a condition from a desired final pose. Ordinary precision and generic proximity alone do not establish this need.

ONLINE EVIDENCE
- Assess object-level progress, incorrect placements, and repeated ineffective attempts. Robot movement alone is not task progress.
- Recognize supported reasoning or constraint-planning needs before a visible failure occurs.
- Use relevant recent views and feedback when assessing recovery or missing information; one occluded view alone does not establish uncertainty.
- Demonstrations belong to a different episode. Use them as procedural references without copying episode-specific identities, counts, or order.
- Public task conditions describe success requirements, not observed hidden outcomes.

USING THE FIXED TASK PRIOR
- Interpret current observations and execution feedback together with the fixed task prior.
- Use the prior to recognize capability requirements that may not be apparent from the next motion alone. A simple-looking action may still belong to an operation requiring reasoning or constraint planning.
- Treat the prior as fallible task-level guidance, not current failure evidence or an executor assignment. It neither overrides current observations nor prevents policy continuation or Astra handback.

CATEGORY DEFINITIONS AND EXECUTOR MAPPING
policy -> Policy:
  Familiar routine manipulation with a suitable observed continuation.
policy_precision -> Policy:
  Fine manipulation without an additional reasoning or process-constraint planning requirement.
astra_semantic -> Astra:
  Task-specific reasoning, retained reference memory, or explicit physical-constraint planning.
astra_correction -> Astra:
  A concrete visible error requiring correction.
astra_recovery -> Astra:
  Repeated ineffective attempts, sustained lack of task progress, or an abnormal state requiring recovery.
astra_uncertain -> Astra:
  Essential task or contact information remains unavailable across relevant observations.

CONTROL PROTOCOL AND OUTPUT
Classify at the controller's allocation gates. Once invoked, Astra owns its intervention and explicitly hands control back; the classifier does not confirm that handback. Return one typed category, without robot actions or task-success claims. AIC applies the category mapping and its allocation rules to select the executor.
\end{lstlisting}

\subsection{One-Time Task-Prior Assessment}
\label{sec:appendix:assessment-prompt}

Astra assesses each episode once at medium reasoning effort, and the resulting prior is reused by all subsequent classifications, which average 49.03 per episode in the RoboDojo main experiment. A single assessment thus amortizes task-level reasoning across recurring allocation decisions.

\begin{lstlisting}[style=appendixprompt]
ROLE
Assess overall task capability requirements once before execution. Produce a task-level prior for later lightweight classification, rather than choosing an executor or proposing actions.

INPUTS
- Task instruction, description, and public success conditions.
- Initial RGB observations and robot state.
- The TDC demonstration reference.

ASSESSMENT PRINCIPLES
Policy competence covers familiar demonstrated manipulation, local alignment, repeated actions, and fixed procedural steps.

routine:
  A supported familiar manipulation procedure, including local alignment, repeated actions, and fixed procedural steps. Duration, object count, and precision alone do not establish additional reasoning needs.
reasoning:
  Task-specific inference, calculation, counting, assignment, ordering, confirmation, or retained reference information beyond a supported fixed procedure.
physical_constraint:
  An explicit condition that must hold during an interaction or intervening motion. A final arrangement or generic collision concern alone is insufficient.

Select routine alone when the task is supported by a familiar procedure. Otherwise, select reasoning and/or physical_constraint only when each is supported by the supplied evidence. When evidence is insufficient, return an empty capability list and mark the assessment accordingly.

Do not infer policy success from its model name or a leaderboard. Treat demonstration content as a procedural reference from another episode. Public success conditions are not observed outcomes. Do not use tools, output actions, claim success, or lock execution to either executor.

STRUCTURED OUTPUT
{
  "assessment_status": "ok | insufficient_evidence",
  "task_capabilities": ["routine | reasoning | physical_constraint"],
  "confidence": "low | medium | high",
  "evidence": ["short statements grounded in the supplied input"]
}

The capability list may contain multiple supported non-routine requirements or be empty when evidence is insufficient. Its values are reusable guidance, not execution evidence or an executor assignment.
\end{lstlisting}

\section{Real-World Experiments}
\label{sec:appendix:real-world}

\subsection{Experimental Details and Results}
\label{sec:appendix:real-world-details}

\begin{figure}[t]
    \centering
    \resizebox{\linewidth}{!}{\input{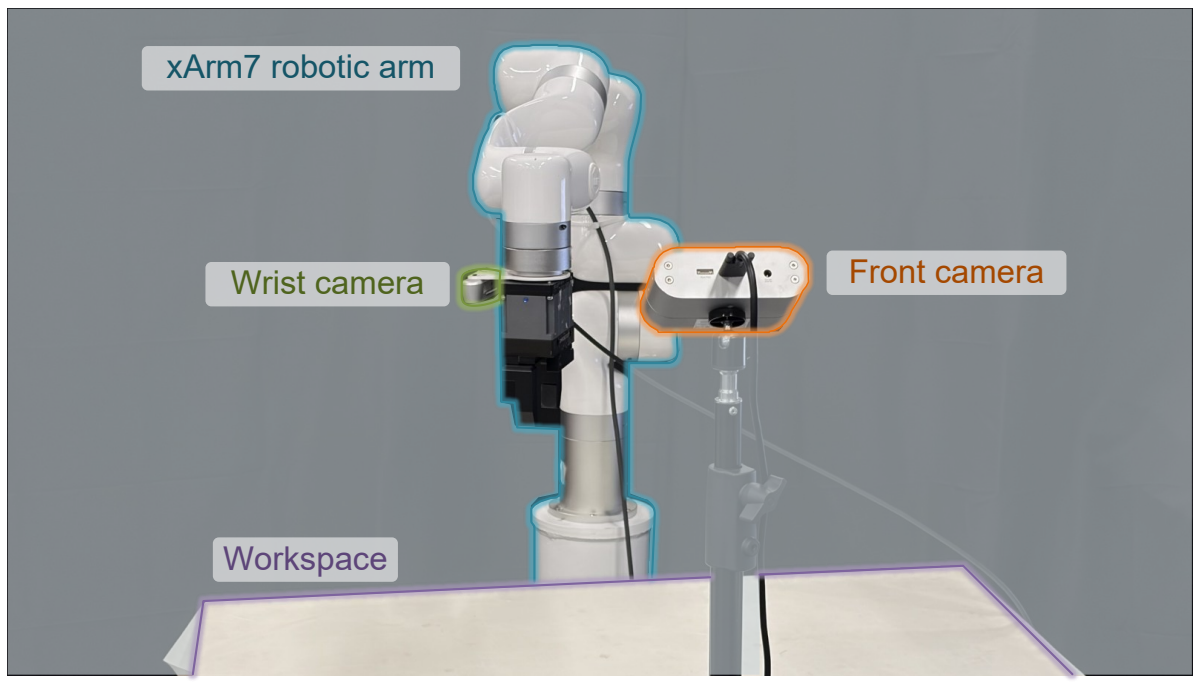}\unskip}
    \captionsetup{skip=10pt}
    \caption{Real-world experimental setup, including the robot, cameras, and workspace.}
    \label{fig:real-robot-setup}
\end{figure}

\paragraph{Robot setup.}
We use an xArm7 robot with a DH Robotics PGI-140-80 gripper, a front-view Orbbec RGB camera, and a wrist-mounted RealSense RGB camera. The manipulation policy is executed by a 20\,Hz joint-position controller, while Astra issues actions in its native action representation (Sec.~\ref{sec:method:setup}). The cameras acquire RGB images at $1920\times1080$ and $640\times480$, respectively, both at 30\,fps. \Cref{fig:real-robot-setup} shows the experimental setup.

\begin{figure}[!t]
    \centering
    \begin{subfigure}[t]{\dimexpr(\linewidth-0.02\linewidth)/2\relax}
        \centering
        \includegraphics[width=\linewidth]{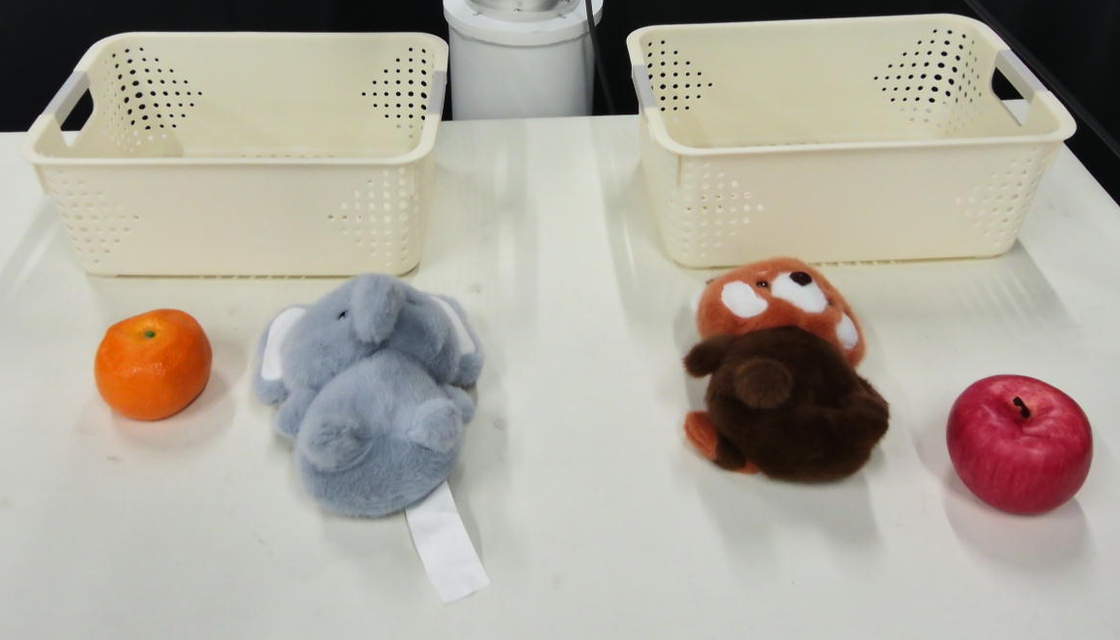}
        \caption{Classify Objects}
        \label{fig:real-robot-classify}
    \end{subfigure}\hspace{0.02\linewidth}%
    \begin{subfigure}[t]{\dimexpr(\linewidth-0.02\linewidth)/2\relax}
        \centering
        \includegraphics[width=\linewidth]{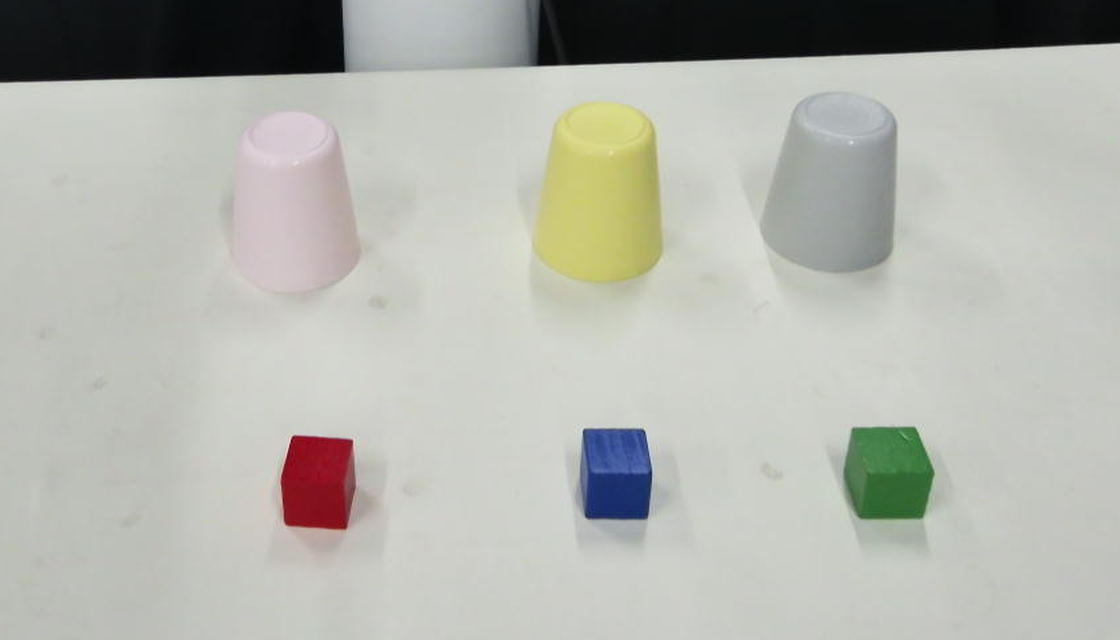}
        \caption{Cover Blocks}
        \label{fig:real-robot-cover}
    \end{subfigure}
    \par\medskip
    \begin{subfigure}[t]{\dimexpr(\linewidth-0.02\linewidth)/2\relax}
        \centering
        \includegraphics[width=\linewidth]{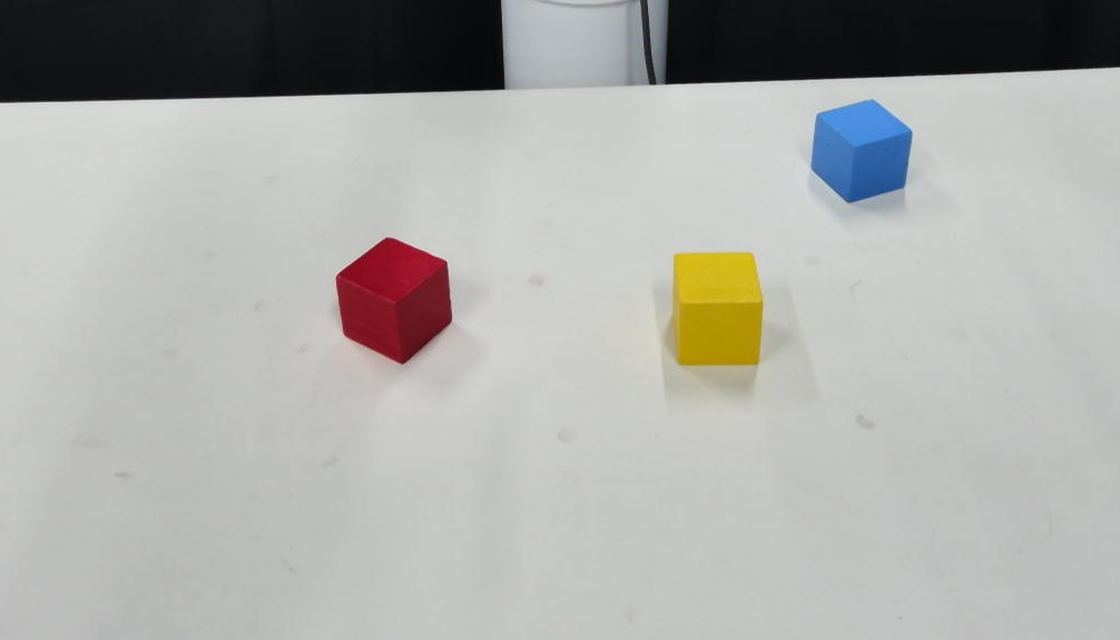}
        \caption{Stack Blocks By Language}
        \label{fig:real-robot-stack}
    \end{subfigure}\hspace{0.02\linewidth}%
    \begin{subfigure}[t]{\dimexpr(\linewidth-0.02\linewidth)/2\relax}
        \centering
        \includegraphics[width=\linewidth]{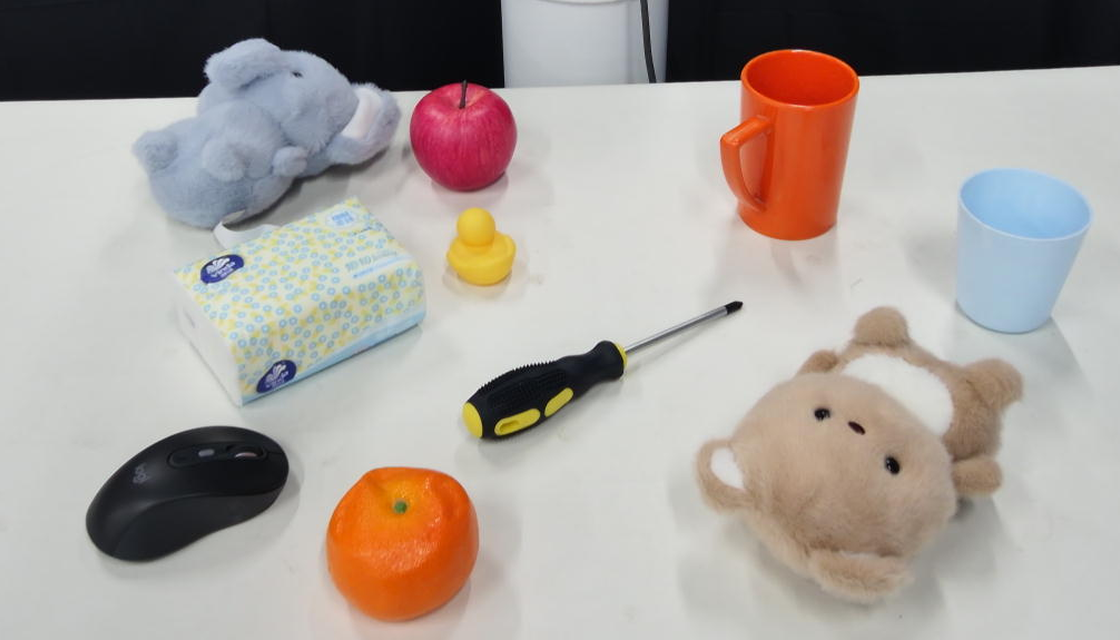}
        \caption{General Pickup}
        \label{fig:real-robot-pickup}
    \end{subfigure}
    \caption{Initial scenes for the four real-world tasks.}
    \label{fig:real-robot-tasks}
\end{figure}

\paragraph{Tasks and instructions.}
We adapt four tabletop tasks from RoboDojo~\cite{chen2026robodojo}; their initial scenes are shown in \cref{fig:real-robot-tasks}.

\textbf{Classify Objects.} The robot places toys in the left basket and fruits in the right basket; evaluation trials vary only the object positions. The instruction is: ``Put toy objects into the left basket and fruit objects into the right basket.''

\textbf{Cover Blocks.} The robot covers three colored blocks with cups from left to right, then uncovers them in red, green, and blue order. It must retain the color--location associations after occlusion. Evaluation trials vary the block color order and cup positions. The instruction is: ``Cover the blocks from left to right, remember their colors, then uncover them in the order: red, green, and blue.''

\textbf{Stack Blocks By Language.} The robot stacks three colored blocks in the order specified by the instruction, requiring color and order understanding as well as precise placement. Evaluation trials vary the instructed color order and block positions. An example instruction is: ``Stack the blocks in the order red, blue, and yellow from bottom to top.''

\textbf{General Pickup.} The robot selects the instructed target from multiple objects and lifts it at least 10\,cm, requiring target identification among distractors and grasping. Evaluation trials vary the target object and object positions. An example instruction is: ``Pick up the spoon by 10\,cm.''

\paragraph{Policy training and model configuration.}
All demonstrations are collected through human teleoperation using a PICO 4 Ultra. For each of the four tasks, $\pi_{0.5}$ is fine-tuned using LoRA~\cite{hu2021lora} on 50 complete demonstrations for 30,000 steps, with a batch size of 32, an initial learning rate of $2.5\times10^{-5}$, and the image encoder frozen. The policy predicts 50 actions at a time; only the first ten are executed. The GPT-6 Astra model identifier is \texttt{gpt-6-astra}, with reasoning effort set to \texttt{medium}. The Agent Invocation Controller uses the lightweight \texttt{gpt-6-luna-decisions} model for online executor selection.

\begin{figure}[!tbp]
    \centering
    \includegraphics[width=0.86\linewidth]{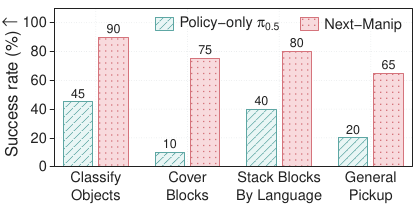}
    \captionsetup{skip=6pt}
    \caption{Real-world success rates (\%) for four tasks. Policy-only $\pi_{0.5}$ and Next-Manip are each evaluated with 20 trials per task.}
    \label{fig:real-world-results}
\end{figure}

\paragraph{Evaluation protocol.}
Next-Manip and the policy-only baseline share the task-specific checkpoint; TDC uses one successful trajectory selected from the same demonstrations, with its layout excluded from evaluation. For each task, both methods are evaluated on the same initial layouts, with model parameters fixed, and success rates are computed from recorded operator labels. Each method is evaluated for 20 trials on each of the four tasks. \Cref{fig:real-world-results} reports the resulting success rates.

\FloatBarrier
\subsection{Supplementary Video}
\label{sec:appendix:real-world-qualitative}

To visualize how agent--policy collaboration unfolds during real-world execution, we provide an accompanying supplementary video with examples from the evaluation rollouts. The video presents a pickup comparison between the policy-only baseline and Next-Manip, examines grasp recovery and control handovers in detail, and shows successful Next-Manip rollouts across all four tasks, with annotations indicating execution stages and key actions. These examples illustrate the design principle of coordinating complementary reasoning and control capabilities on demand, adapting their roles to changing execution needs.

\section{Executor Allocation Statistics}
\label{sec:appendix:allocation}

Table~\ref{tab:executor-allocation} summarizes executor allocation on the 12 representative RoboDojo tasks, with 20 episodes per task. Tasks whose priors indicate reasoning demands (\emph{Largest}, \emph{Swap}, and \emph{Conveyor}) are executed almost entirely by Astra, with 95.39\% to 97.19\% of the steps, whereas tasks with routine priors are executed mainly by the policy. On these reasoning tasks, Astra retains control across multiple action chunks once allocated (Sec.~\ref{sec:method:invocation}), with only 2.10 to 3.85 classifier-driven allocations to Astra per episode. Policy stalls also trigger agent takeovers, averaging 0.30 per episode, which account for Astra's execution on tasks with no classifier-driven allocation to Astra. These allocation patterns show that AIC, guided by the task prior, broadly matches executors to task demands, bringing the complementary strengths of the agent and the policy to bear where they are needed, consistent with the analysis in Sec.~\ref{sec:experiments:analysis}.

\begin{table*}[!t]
  \centering
  \caption{Executor allocation on 12 representative RoboDojo tasks, with 20 episodes per task. Task priors are identical across episodes within each task. Overall reports the mean over tasks.}
  \label{tab:executor-allocation}
  \footnotesize
  \setlength{\tabcolsep}{5pt}
  \renewcommand{\arraystretch}{1.15}
  \begin{tabular*}{\textwidth}{@{\extracolsep{\fill}}lllrrrr@{}}
    \toprule
    \textbf{\raisebox{-3pt}[0pt][0pt]{Category}} & \textbf{\raisebox{-3pt}[0pt][0pt]{Task}} & \textbf{\raisebox{-3pt}[0pt][0pt]{Task prior}} & \multicolumn{2}{c}{\textbf{\shortstack{Classifier-driven allocations\\per episode}}} & \textbf{\raisebox{-8pt}[0pt][0pt]{\shortstack{Stall takeovers\\per episode}}} & \textbf{\raisebox{-6.4pt}[0pt][0pt]{\shortstack{Astra execution\\step share (\%)}}} \\
    \cmidrule(lr){4-5}
     & & & Policy & Astra & & \\
    \midrule
     & Mugs & Routine & 75.00 & 1.05 & 0.25 & 14.39 \\
    Generalization & Pack & Routine & 111.40 & 0.15 & 1.40 & 15.13 \\
     & Largest & Reasoning & 0.05 & 3.85 & 0.00 & 95.39 \\
    \addlinespace[3pt]
     & Charger & Routine & 42.75 & 0.00 & 0.35 & 7.11 \\
    Precision & Tubes & Routine & 42.00 & 0.00 & 0.25 & 6.66 \\
     & Tower & Routine & 85.50 & 0.10 & 0.00 & 1.98 \\
    \addlinespace[3pt]
     & Kong & Routine & 14.65 & 19.10 & 0.00 & 40.11 \\
    Long-Horizon & Egg & Routine & 64.15 & 1.95 & 0.35 & 21.26 \\
     & Organize & Routine & 76.90 & 0.15 & 1.00 & 19.20 \\
    \addlinespace[3pt]
     & Swap & Reasoning & 0.10 & 2.10 & 0.00 & 97.19 \\
    Memory & Cover & Routine & 6.05 & 38.90 & 0.00 & 18.12 \\
     & Conveyor & Reasoning & 0.00 & 2.40 & 0.00 & 96.40 \\
    \midrule
    Overall & --- & --- & 43.21 & 5.81 & 0.30 & 36.08 \\
    \bottomrule
  \end{tabular*}
  \par\vspace{12.93pt}
\end{table*}

\begin{figure*}[!t]
  \centering
  \includegraphics[width=\textwidth]{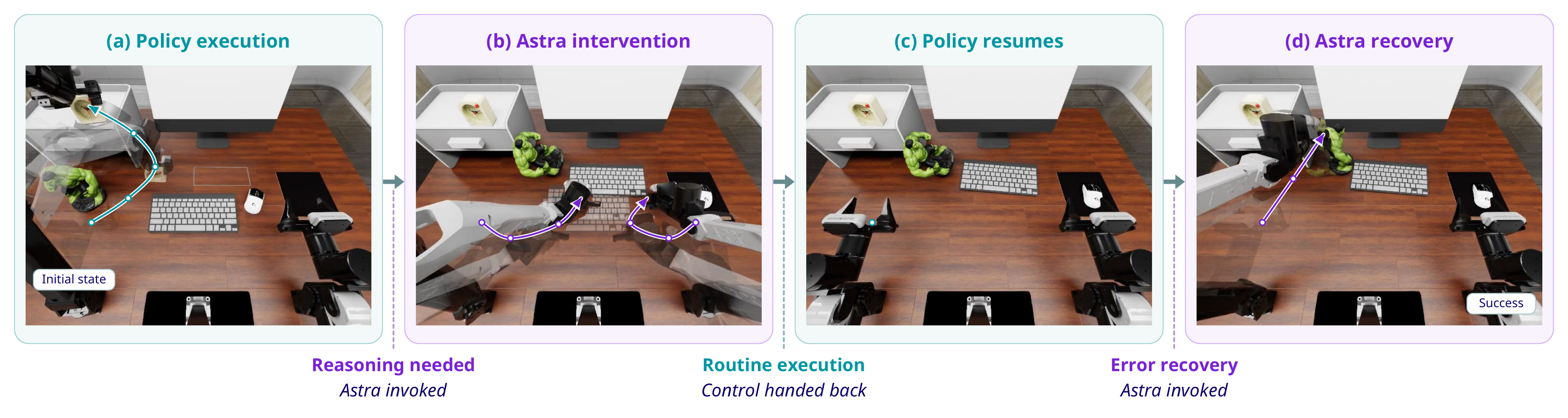}
  \captionsetup{skip=2pt}
  \caption{Illustrative collaboration on RoboDojo's Organize Table task. Trajectories visualize execution by the policy and Astra, including control handback and renewed intervention within one episode.}
  \label{fig:organize-collaboration}
\end{figure*}

\section{Qualitative Analysis of Control Handover}
\label{sec:appendix:qualitative}

To complement the aggregate results, we examine how the two executors advance the task within an episode. In the \emph{Organize Table} example in \cref{fig:organize-collaboration}, the policy first manipulates the alarm clock. When local execution does not advance the remaining subgoals, Astra takes over to reassess task progress and plan the keyboard manipulation, then hands control back to the policy. When policy execution later stalls, Astra intervenes again to correct the figurine placement and complete the task. This sequence illustrates how routine manipulation, reasoning about unfinished subgoals, and error recovery can be coordinated as task demands evolve.

\end{document}